\documentclass[11pt,letterpaper]{article}

\usepackage[letterpaper,total={6.5in,8.5in}]{geometry}
\usepackage[toc,page]{appendix}
\usepackage{graphicx}
\graphicspath{{./fig/}}
\DeclareGraphicsExtensions{.eps}
\usepackage{xcolor,epsfig}
\usepackage[noadjust,sort,compress]{cite}
\usepackage[hyphens]{url}
\usepackage[cmex10]{amsmath}
\usepackage{amsthm,amssymb,bm,mathtools}
\usepackage{amsfonts,mathrsfs}
\usepackage{algorithm,algorithmicx,algpseudocode}
\usepackage[font=normalsize]{subfig}
\usepackage{float}
\usepackage{etoolbox}

\DeclarePairedDelimiterX\norm[1]\lVert\rVert{\ifblank{#1}{\:\cdot\:}{#1}}
\DeclarePairedDelimiterX\innerp[2]{\langle}{\rangle}{#1
  \mathop{}\delimsize|\mathop{} #2}
\DeclareMathOperator{\linspan}{span}
\DeclareMathOperator*{\Argmin}{argmin}
\DeclareMathOperator*{\Argmax}{argmax}
\DeclareMathOperator{\prob}{Pr}

\newcommand\distto{{\mathop{}\|\mathop{}}}
\newcommand{\hilbert}{\mathcal{H}}
\newcommand{\Real}{\mathbb{R}}

\theoremstyle{definition}
\newtheorem{remark}{Remark}

\title{{Sketch and Validate for Big Data Clustering}}

\author{Panagiotis A.~Traganitis,
  Konstantinos~Slavakis,\\
  Georgios~B.~Giannakis%
  \thanks{The authors are with the Dept.\ of Electrical and Computer
    Engineering and
    the Digital Technology Center, at Univ.\ of Minnesota,
    117 Pleasant St.\ SE, Minneapolis, MN 55455, USA. Tel:
    (612)\,625-4287; Emails:
    \{traga003,kslavaki,georgios\}@umn.edu}
  \thanks{Work was supported by NSF grants 1343248, 1343860, 1442686, and
     1500713. Preliminary parts of this work appeared at the
     \emph{Proc.\ of Asilomar Conf.\ on Signals, Systems, and
       Computers}, Pacific Grove, CA, Nov.\ 2014.  
  }
}

\date{\vspace{-5ex}}

\begin{document}
\maketitle

\begin{abstract}
  In response to the need for learning tools tuned to big data
  analytics, the present paper introduces a framework for efficient
  clustering of huge sets of (possibly high-dimensional)
  data. Building on random sampling and consensus (RANSAC) ideas
  pursued earlier in a different (computer vision) context for robust
  regression, a suite of novel dimensionality- and set-reduction
  algorithms is developed. The advocated \textit{sketch-and-validate}
  (SkeVa) family includes two algorithms that rely on
  $K$-means clustering per iteration on reduced number of dimensions
  and/or feature vectors: The first operates in a batch fashion, while
  the second sequential one offers computational efficiency and
  suitability with streaming modes of operation. For clustering even
  nonlinearly separable vectors, the SkeVa family offers also a member
  based on user-selected kernel functions.  Further trading off
  performance for reduced complexity, a fourth member of the SkeVa
  family is based on a divergence criterion for selecting proper
  minimal subsets of feature variables and vectors, thus bypassing the
  need for $K$-means clustering per iteration. Extensive numerical
  tests on synthetic and real data sets highlight the potential of the
  proposed algorithms, and demonstrate their competitive performance
  relative to state-of-the-art random projection alternatives.
  
\end{abstract}


\section{Introduction}\label{sec:intro}

As huge amounts of data are collected perpetually from communication,
imaging, and mobile devices, medical and e-commerce platforms as well
as social-networking sites, this is undoubtedly an era of data
deluge~\cite{bigdata_economist}. Such ``big data'' however, come with
``big challenges.''  The sheer volume and dimensionality of data make
it often impossible to run analytics and traditional inference methods
using stand-alone processors, e.g., \cite{bickel.curse.08,
  MJ.review.bernoulli.13}. Consequently, ``workhorse'' learning tools
have to be re-examined in the face of today's \emph{high-cardinality}
sets possibly comprising \emph{high-dimensional} data.

Clustering (a.k.a.\ unsupervised classification) refers to
categorizing into groups unlabeled data encountered in the widespread
applications of mining, compression, and learning tasks~\cite{bishop}.
Among numerous clustering algorithms, $K$-means is the most prominent
one thanks to its simplicity~\cite{bishop}. It thrives on ``tight''
groups of feature vectors, data points or objects that can be
separated via (hyper)planes. Its scope is further broadened by the
so-termed probabilistic and kernel $K$-means, with an instantiation of
the latter being equivalent to spectral clustering -- the popular tool
for graph-partitioning that can cope even with nonlinearly separable
data~\cite{dhillon2004kernel}.

A key question with regards to clustering data sets of cardinality $N$
containing $D$-dimensional vectors with $N$ and/or $D$ huge, is: How
can one select the most ``informative'' vectors and/or dimensions so
as to reduce their number for efficient computations, yet retain as
much of their cluster-discrimination capability? This paper develops
such an approach for big data $K$-means clustering. Albeit distinct,
the inspiration comes from random sampling and consensus (RANSAC)
methods, which were introduced for outlier-resilient regression tasks
encountered in computer vision~\cite{RANSAC81, torr2000mlesac,
  nister2005preemptive, MultiRANSAC, toldo2008robust,
  chum2008optimal}.

Feature selection is a rich topic \cite{guyon.elisseeff.03} explored
extensively from various angles, including pattern recognition, source
coding and information theory, (combinatorial) optimization
\cite{Zhang.Sun.02, Yu.Yuan.93}, and neural networks
\cite{Chatterjee.97}. Unfortunately, most available selection schemes
do not scale well with the number of features $D$, particularly in the
big data regime where $D$ is massive. Recent approaches to
dimensionality reduction and clustering include subspace clustering,
where minimization problems requiring singular value decompositions
(SVDs) are solved per iteration to determine in parallel a
low-dimensional latent space and corresponding sparse coefficients for
efficient clustering~\cite{patel.iccv.13}. Low-dimensional subspace
representations are also pursued in the context of kernel $K$-means
\cite[Alg.~2]{chitta.kdd.11}, where either an SVD on a sub-sampled kernel
matrix, or, the solution of a quadratic optimization task is required
per iteration to cluster efficiently large-scale data.

Randomized schemes select features with non-uniform
  probabilities that depend on the so-termed ``leverage scores'' of
  the data matrix~\cite{mahoney2011randomized,
    mahoney2012algorithmic}. Unfortunately, their complexity is loaded
  by the leverage scores computation, which, similar
  to~\cite{patel.iccv.13}, requires SVD computations -- a cumbersome
  (if not impossible) task when $D\gg$ and/or $N\gg$. Recent
  computationally efficient alternatives for feature selection and
  clustering rely on random projections (RPs)
  \cite{mahoney2011randomized, mahoney2012algorithmic,
    clarkson.woodruff.stoc.13, Drineas}. Specifically for RP-based
  clustering~\cite{Drineas}, the data matrix is left multiplied by a
  data-agnostic (fat) $d\times D$ RP matrix to reduce its dimension
  ($d\ll D$); see also \cite{kaist.kernel.kmeans.nips.11} where RPs
  are employed to accelerate the kernel $K$-means
  algorithm. Clustering is performed afterwards on the reduced
  $d$-dimensional vectors. With its universality and quantifiable
  performance granted, this ``one-RP-matrix-fits-all'' approach is not
  as flexible to adapt to the data-specific attributes (e.g., low
  rank) that is typically present in big data.
  
This paper's approach aspires to not only account for structure, but
also offer a gamut of novel randomized algorithms trading off
complexity for clustering performance by developing a family of what
we term sketching and validation (SkeVa) algorithms. The SkeVa family
includes two algorithms based on efficient intermediate $K$-means
clustering steps. The first is a batch method, while the second is
sequential thus offering computational efficiency and suitability for
streaming modes of operation. The third member of the SkeVa family is
kernel-based and enables big data clustering of even nonlinearly
separable data sets. Finally, the fourth one bypasses the need for
intermediate $K$-means clustering thus trading off performance for
complexity. Extensive numerical validations on synthetic and real
data-sets highlight the potential of the proposed methods, and
demonstrate their competitive performance on clustering massive
populations of high-dimensional data vs.\ state-of-the-art RP
alternatives.

\noindent\textbf{Notation.} Boldface uppercase (lowercase) letters indicate
matrices (column vectors). Calligraphic uppercase letters denote sets
($\varnothing$ stands for the empty set), and $|\mathcal{A}|$
expresses the cardinality of $\mathcal{A}$. Operators $\|\cdot\|_2$
and $\|\cdot\|_1$ stand for the $\ell_2$- and $\ell_1$-norm of a
vector, respectively, while $(\cdot)^{\top}$ denotes transposition and
$\bm{1}$ the all-one vector.

\section{Preliminaries}\label{sec:prelim}

Consider the $D\times N$ data matrix $\bm{X} :=
[\bm{x}_1,\ldots,\bm{x}_N]$ with $D$ and/or $N$ being potentially
massive. Data $\{\bm{x}_n\}_{n=1}^{N}$ belong to a known number of $K$
clusters ($K\ll N$). Each cluster $\mathcal{C}_k$ is represented by
its centroid $\bm{c}_k$ that can be e.g., the (sample) mean of the
vectors in $\mathcal{C}_k$. Accordingly, each datum can be modeled as
$\bm{x}_n = \bm{C}\bm{\pi}_n + \bm{v}_n$, where $\bm{C}:= [\bm{c}_1,
\ldots, \bm{c}_K]$; the sparse $K\times 1$ vector $\bm{\pi}_n$
comprises the data-cluster association entries satisfying
$\sum_{k=1}^K [\bm{\pi}_n]_k =1$, where $[\bm{\pi}_n]_k \in (0,1]$
when $\bm{x}_n\in \mathcal{C}_k$, while $[\bm{\pi}_n]_k =0$,
otherwise; and the noise $\bm{v}_n$ captures $\bm{x}_n$'s deviation
from the corresponding centroid(s).

For \emph{hard clustering}, the said associations are binary
($[\bm{\pi}_n]_k\in\{0,1\}$), and in the celebrated hard $K$-means
algorithm they are identified based on the Euclidean ($\ell_2$)
distance between $\bm{x}_n$ and its closest centroid. Specifically,
given $K$ and $\{\bm{x}_n\}_{n=1}^{N}$, per iteration $i=1,2,\ldots$,
the $K$-means algorithm iteratively updates data-cluster associations
and cluster centroids as follows; see e.g., \cite{bishop}.
\begin{subequations}\label{kmeans}
  \begin{align}
    \intertext{[$i$-a] \textbf{Update data-cluster associations:} For
      $n=1,\ldots,N$,} \bm{x}_n\in\mathcal{C}_k[i] & \Leftrightarrow k
    \in \Argmin_{k'\in\{1,
      \ldots, K\}} \norm*{\bm{x}_n-\bm{c}_{k'}[i]}_2^2 \;.\label{kmeans.a}\\
    \intertext{[$i$-b] \textbf{Update cluster centroids:} For
      $k=1,\ldots,K$,} \bm{c}_k[i+1] & \in \Argmin_{\bm{c}\in\Real^D}
    \sum_{\bm{x}_n\in \mathcal{C}_k[i]}
    \norm{\bm{x}_n - \bm{c}}_2^2  \notag\\
    & = \frac{1}{\bigl|\mathcal{C}_k[i]\bigr|} \sum\nolimits_{\bm{x}_n
      \in \mathcal{C}_k[i]} \bm{x}_n \;.\label{kmeans.b}
  \end{align}
\end{subequations}
Although there may be multiple assignments solving \eqref{kmeans.a},
each $\bm{x}_n$ is assigned to a single cluster. To initialize
\eqref{kmeans.a}, one can randomly pick $\{\bm{c}_k[1]\}_{k=1}^K$ from
$\{\bm{x}_n\}_{n=1}^{N}$. The iterative algorithm \eqref{kmeans}
solves a challenging NP-hard problem, and albeit its success,
$K$-means guarantees convergence only to a local minimum at complexity
$\mathcal{O}(NDKI)$, with $I$ denoting the number of iterations needed
for convergence, which depends on
initialization~\cite[\S~9.1]{bishop}.

\begin{remark}\label{rem:kmedoids}
  As \eqref{kmeans.a} and \eqref{kmeans.b} minimize an $\ell_2$-norm
  squared, hard $K$-means is sensitive to outliers. Variants
  exhibiting robustness to outliers adopt non-Euclidean distances
  (a.k.a. dissimilarity metrics) $\delta$, such as the
  $\ell_1$-norm. In addition, candidate ``centroids'' can be selected
  per iteration from the data themselves; that is, $\bm{c}\in
  \mathcal{C}_k[i]$ in \eqref{kmeans.b}. Notwithstanding for this
  so-termed \textit{$K$-medoids} algorithm, one just needs the
  distances $\{\delta (\bm{x}_n,\bm{x}_{n'})\}$ to carry out the
  minimizations in \eqref{kmeans.a} and \eqref{kmeans.b}. The latter
  in turn allows $\{\bm{x}_n\}_{n=1}^{N}$ to even represent
  \emph{non-vectorial} objects (a.k.a. qualitative data), so long as
  the aforementioned (non-)Euclidean distances can become available
  otherwise; e.g., in the form of correlations~\cite[\S~9.1]{bishop}.
\end{remark}

Besides various distances and centroid options, hard $K$-means in
\eqref{kmeans} can be generalized in three additional directions: (i)
Using nonlinear functions $\varphi:\Real^D\to\hilbert$, with
$\hilbert$ being a potentially infinite-dimensional space, data
$\{\bm{x}_n\}$ can be transformed to $\{\varphi(\bm{x}_n)\}$, where
clustering can be facilitated (cf.\ Sec.~\ref{sec:kernel.version});
(ii) via non-binary $\bm{\pi}_n\in [0,1]^K$, \emph{soft clustering}
can allow for multiple associations per datum, and thus for a
probabilistic assignment of data to clusters; and (iii) additional
constraints (e.g., sparsity) can be incorporated in the
$[\bm{\pi}_n]_k$ coefficients through appropriate regularizers
$\rho(\bm{\pi})$.

All these generalizations can be unified by
replacing \eqref{kmeans} per iteration $i=1,2,\ldots$, with
\begin{subequations}\label{soft.kmeans}
  \begin{align}
    \intertext{[$i$-a] \textbf{Update data-cluster associations:}\
      $n=1,\ldots,N$,}
    & \bm{\pi}_n[i] \in \Argmin_{\substack{\bm{\pi}\in [0,1]^K;\\
        \bm{1}^{\top}\bm{\pi}=1}}
    \delta\left(\varphi(\bm{x}_n),
      \sum_{k=1}^K\pi_k \bm{c}_k[i]\right) +
    \rho(\bm{\pi})\,.\label{soft.kmeans.a}\\
    \intertext{[$i$-b] \textbf{Update cluster centroids:}}
    & \{\bm{c}_k[i+1]\}_{k=1}^K \in
    \smashoperator[l]{\Argmin_{\{\bm{c}_ k \}_{k=1}^K
        \subset\hilbert}}
    \sum_{n=1}^N \delta\left(\varphi(\bm{x}_n),\sum_{k=1}^K
      \bigl[\bm{\pi}_n[i]\bigr]_k \bm{c}_k
    \right)\,. \label{soft.kmeans.b}
  \end{align}
\end{subequations}
In Sec.~\ref{sec:kernel.version}, function $\varphi$ will be
implicitly used to map nonlinearly separable data
$\{\bm{x}_n\}_{n=1}^N$ to linearly separable (possibly infinite
dimensional) data $\{\varphi(\bm{x}_n)\}_{n=1}^N$, whose distances can
be obtained through a pre-selected (so-termed kernel) function
$\kappa$~\cite[Chap.~6]{bishop}. The regularizer $\rho(\bm{\pi})$ can
enforce prior knowledge on the data-cluster association vectors.

To confirm that indeed \eqref{kmeans} is subsumed by
\eqref{soft.kmeans}, let $\varphi(\bm{x}_n) = \bm{x}_n$; choose
$\delta$ as the squared Euclidean distance in $\Real^D$; and set $\rho
(\bm{\pi}) = 0$, if $\bm{\pi} \in \mathcal{E}_K :=
\{\bm{e}_k\}_{k=1}^K$, whereas $\rho (\bm{\pi}) = +\infty$ otherwise,
with $\bm{e}_k$ being the $k$th $K$-dimensional canonical vector. Then
\eqref{soft.kmeans.a} becomes $\bm{\pi}_n[i]\in \Argmin_{\bm{\pi}\in
  \{0,1\}^K;\,\bm{1}^{\top}\bm{\pi}=1} \norm{\bm{x}_n - \sum_k\pi_k
  \bm{c}_k[i]}_2^2$, which further simplifies to the minimum distance
rule of \eqref{kmeans.a}. Moreover, it can be readily verified that
\eqref{soft.kmeans.b} $\{\bm{c}_k[i+1]\}_{k=1}^K \in
\Argmin_{\{\bm{c}_k\}_{k=1}^K} \sum_n\norm{\bm{x}_n - \sum_k
  [\bm{\pi}_n[i]]_k \bm{c}_k}_2^2$ separates across $k$s to yield the
centroid of \eqref{kmeans.b} per cluster.

To recognize how \eqref{soft.kmeans} captures also soft clustering,
consider that cluster $\mathcal{C}_k$ is selected with probability
$\pi_k:=\prob(\mathcal{C}_k)$, and its data are drawn from a
probability density function (pdf) $p$ parameterized by
$\bm{\theta}_k$; i.e., $\bm{x}|\mathcal{C}_k\sim p(\bm{x};
\bm{\theta}_k)$.  If $p$ is Gaussian, then $\bm{\theta}_k$ denotes its
mean $\bm{\mu}_k$ and covariance matrix $\bm{\Sigma}_k$. With
$\bm{\theta}:= [\bm{\theta}_1^{\top}, \ldots,
\bm{\theta}_K^{\top}]^{\top}$ and allowing for multiple cluster
associations, the likelihood per datum is given by the mixture pdf: $p
(\bm{x};\bm{\pi},\bm{\theta}) = \sum_{k=1}^K \pi_k p(\bm{x} ;
\bm{\theta}_k)$, which for independently drawn data yields the joint
log-likelihood ($\bm{\Pi}:= [\bm{\pi}_1, \ldots, \bm{\pi}_N]$)
\begin{equation}
  \ln\,p(\bm{X};\bm{\Pi},\bm{\theta}) = \sum_{n=1}^N\ln \biggl(\sum_{k=1}^K
  [\bm{\pi}_n]_k p(\bm{x}_n;\bm{\theta}_k)\biggr)\,. \label{log.likelihood}
\end{equation}
If $\varphi(\bm{x}_n):=\bm{x}_n$, $\bm{c}_k :=
p(\bm{x}_n;\bm{\theta}_k)$, and $\delta(\bm{x}_n, \sum_k\pi_kc_k):= -
\ln (\sum_k [\bm{\pi}_n]_k p(\bm{x}_n;\bm{\theta}_k))$ in
\eqref{soft.kmeans}, then soft $K$-means iterations
\eqref{soft.kmeans.a} and \eqref{soft.kmeans.b} maximize
\eqref{log.likelihood} with respect to (w.r.t.) $\bm{\Pi}$ and
$\bm{\theta}$. An alternative popular maximizer of the likelihood in
\eqref{log.likelihood} is the \textit{expectation-maximization}
algorithm; see e.g.,~\cite[\S~9.3]{bishop}.

If the number of clusters $K$ is unknown, it can also be estimated by
regularizing the log-likelihood in \eqref{log.likelihood} with terms
penalizing complexity as in e.g., minimum description length
criteria~\cite{bishop}.

Although hard $K$-means is the clustering module common to all
numerical tests in Sec. V, the novel big data algorithms of Secs. III
and IV apply to all schemes subsumed by \eqref{soft.kmeans}.

\section{The SkeVa Family}

Our novel algorithms based on random sketching and validation are
introduced in this section. Relative to existing clustering schemes,
their merits are pronounced when $D$ and/or $N$ take prohibitively
large values for the unified iterations \eqref{soft.kmeans.a} and
\eqref{soft.kmeans.b} to remain computationally feasible.

\subsection{Batch algorithm}\label{sec:algo}

For specificity, the SkeVa~K-means algorithm will be developed first for
$D\gg$, followed by its variant for $N\gg$.

{Using a repeated trial-and-error approach,
  SkeVa~K-means discovers a few dimensions (features) that yield
  high-accuracy clustering. The key idea is that upon sketching a
  small enough subset of dimensions (trial or sketching phase), a
  hypotheses test can be formed by augmenting the original subset with
  a second small subset (up to $d$ affordable dimensions) to validate
  whether the first subset nominally represents the full
  $D$-dimensional data (error phase). Such a trial-and-error procedure
  is repeated for a number $R_{\max}$ of realizations, after which the
  features that have achieved the ``best'' clustering accuracy results
  are those determining the final clusters on the whole set of
  dimensions.}

Starting with the trial-phase per realization $r$, $\check{d}$
dimensions (rows) of $\bm{X}$ are randomly drawn (uniformly) to obtain
$\check{\bm{X}}^{(r)}:= [\check{\bm{x}}_1^{(r)},
\ldots,\check{\bm{x}}_N^{(r)}] \in \Real^{\check{d}\times N}$. With
$\check{d}$ small enough, $K$-means is run on $\check{\bm{X}}^{(r)}$
to obtain clusters $\{\check{\mathcal{C}}_k^{(r)}\}_{k=1}^K$ and
corresponding centroids $\{\check{\bm{c}}_k^{(r)}\}_{k=1}^K$ [cf.\
\eqref{kmeans.a} and \eqref{kmeans.b}]. These sketching and clustering
steps comprise the \textit{(random) sketching} phase.

Moving on to the error-phase of the procedure, the quality of the
$\check{d}$-dimensional clustering is assessed next using what we term
\textit{validation} phase. This starts by re-drawing
$\check{d}'$-dimensional data $\{\check{\bm{x}}_n^{(r')}\}_{n=1}^N$
$(\check{d}' \ll D-\check{d})$, generally different from those
selected in draw $r$. Associating each $\check{\bm{x}}_n^{(r')}$ with
the cluster $\check{\bm{x}}_n^{(r)}$ belongs to, the centroids
corresponding to the extra $\check{d}'$ dimensions are formed as
[cf.~\eqref{kmeans.b}]
\begin{equation}
\check{\bm{c}}_k^{(r')} = \frac{1}{\bigl\lvert\check{\mathcal{C}}_k^{(r)}
  \bigr\rvert} \sum_{\check{\bm{x}}_n^{(r)}\in\check{\mathcal{C}}_k^{(r)}}
\check{\bm{x}}_n^{(r')}\,. \label{centroid.extra.dims}
\end{equation}
Let $\bar{\bm{x}}_n^{(r)} := [\check{\bm{x}}_n^{(r)}{}^{\top},
\check{\bm{x}}_n^{(r')}{}^{\top}]^{\top}$ and $\bar{\bm{c}}_k^{(r)} :=
[\check{\bm{c}}_k^{(r)}{}^{\top},\check{\bm{c}}_k^{(r')}{}^{\top}]^{\top}$
denote respectively the concatenated data and centroids from draws $r$
and $r'$, and likewise for the data and centroid matrices
$\bar{\bm{X}}^{(r)}$ and $\bar{\bm{C}}^{(r)}$. Measuring distances
$\{\delta (\bar{\bm{x}}_n^{(r)},\bar{\bm{c}}_k^{(r)})\}$ and using
again the minimum distance rule data-cluster associations and clusters
$\{\bar{\mathcal{C}}_k^{(r)}\}_{k=1}^K$ are obtained for the
``augmented data.''  If per datum $\bm{x}_n$ the data-cluster
association in the space of $\check{d}$ dimensions coincides with that
in the space of $d:= \check{d} + \check{d}'$ dimensions, then
$\bm{x}_n$ is in the \textit{validation set} (VS)
$\mathcal{V}_D^{(r)}$; that is,
\begin{equation}
  \mathcal{V}_D^{(r)} := \left\{\bm{x}_n \Bigm\vert \check{\bm{x}}_n^{(r)} \in
    \check{\mathcal{C}}_{k_1}^{(r)}, \bar{\bm{x}}_n^{(r)} \in
    \bar{\mathcal{C}}_{k_2}^{(r)},\ \text{and}\ k_1=k_2 \right\}\,.\label{CS}
\end{equation}
Quality of clustering per draw is then assessed using a monotonically
increasing rank function $f$ of the set $\mathcal{V}_D^{(r)}$. Based
on this function, a $\check{d}$-dimensional trial $r_1$ is preferred
over another $\check{d}$-dimensional trial $r_2$ if
$f(\mathcal{V}_D^{(r_1)}) > f(\mathcal{V}_D^{(r_2)})$.

The sketching and validation phases are repeated for a prescribed
number of realizations $R_{\max}$. At last, the
$\check{d}$-dimensional sketching $r_* :=
\Argmax_{r\in\{1,\ldots,R_{\max}\}} f(\mathcal{V}_D^{(r)})$ yields the
final clusters, namely $\{\check{\mathcal{C}}_k^{(r_*)}\}_{k=1}^K$;
see Alg.~\ref{alg:SkeVa}.

\begin{algorithm}
  \begin{algorithmic}[1]
    \algrenewcommand\algorithmicindent{1em}
    \Require{Data $\bm{X}$; number of clusters $K$; reduced dimensions
      $\check{d}$ and $\check{d}'$ for the sketching and
      validation phases, respectively; ranking function $f$; number of
      realizations (draws) $R_{\max}$.}
    \Ensure{Data-cluster associations on $\bm{X}$.}
    \For{$r = 1$ to $R_{\max}$}
    \State\parbox[t]{\dimexpr\linewidth-\algorithmicindent}{Randomly sample
      $\check{d} \ll D$ rows of $\bm{X}$ to obtain
      $\check{\bm{X}}^{(r)}$.}\label{alg1.step:random.sample}
    \State\parbox[t]{\dimexpr\linewidth-\algorithmicindent}{Run $K$-means on
      $\check{\bm{X}}^{(r)}$; obtain clusters
      $\{\check{\mathcal{C}}_k^{(r)}\}_{k=1}^K$ and centroids
      $\{\check{\bm{c}}_k^{(r)}\}_{k=1}^K$ [cf.\
      \eqref{kmeans}].}\label{alg1.step:kmeans}
    \State\parbox[t]{\dimexpr\linewidth-\algorithmicindent}{Randomly sample
      $\check{d}' \ll D$ rows of $\bm{X}$ (other than those in
      step~\ref{alg1.step:random.sample}) to obtain $\check{\bm{X}}^{(r')}$.}
    \State\parbox[t]{\dimexpr\linewidth-\algorithmicindent}{Per cluster $k$,
      form $\bar{\bm{c}}_k^{(r)}:= [\check{\bm{c}}_k^{(r)}{}^{\top},
      \check{\bm{c}}_k^{(r')}{}^{\top}]^{\top}$.}
    \State\parbox[t]{\dimexpr\linewidth-\algorithmicindent}{Associate
      $\{\bar{\bm{x}}_n^{(r)}\}_{n=1}^N$ to closest
      $\{\bar{\bm{c}}_k^{(r)}\}_{k=1}^K$.}\label{alg1.step:data.cluster.aug}
    \State\parbox[t]{\dimexpr\linewidth-\algorithmicindent}{Identify
      validation set $\mathcal{V}_D^{(r)}$ [cf.\ \eqref{CS}].}
    \EndFor
    \State{$r_* := \arg\max_{r\in\{1, \ldots, R_{\max}\}}
      f(\mathcal{V}_D^{(r)})$.}
    \State{Associate data to clusters on $\bm{X}$ as in
      $\{\check{\mathcal{C}}_k^{(r_*)}\}_{k=1}^K$.}
  \end{algorithmic}
  \caption{{Batch SkeVa~K-means}}\label{alg:SkeVa}
\end{algorithm}

With regards to selecting $f$, a straightforward choice is the VS
cardinality, that is $f(\mathcal{V}_D^{(r)}) := \lvert
\mathcal{V}_D^{(r)}\rvert$, which can be thought as the empirical
probability of correct clustering. Alternatively, a measure of cluster
separability, used extensively in pattern recognition, is
\textit{Fisher's discriminant ratio} \cite{bishop}, which in the
present context becomes
\begin{equation}\label{eq:fFDRours}
  \text{FDR}^{(r)} := \sum_{k_1=1}^{K}\sum_{\substack{k_2=1;\\ k_2 \neq
      k_1}}^{K}\frac{\norm*{\bar{\bm{c}}_{k_1}^{(r)} -
      \bar{\bm{c}}_{k_2}^{(r)}}_2^2}{\bigl(\bar{\sigma}_{k_1}^{(r)}\bigr)^2 +
    \bigl(\bar{\sigma}_{k_2}^{(r)}\bigr)^2}
\end{equation}
where $(\bar{\sigma}_k^{(r)})^2$ is the unbiased sample variance of
cluster $k$:
\begin{equation}\label{eq:samplevar}
  \bigl(\bar{\sigma}_k^{(r)} \bigl)^2 :=
  \frac{1}{\lvert\check{\mathcal{C}}_k^{(r)} \rvert -1}
  \sum_{\check{\bm{x}}_n
    \in \check{\mathcal{C}}_k^{(r)}} \norm*{\bar{\bm{x}}_n^{(r)} -
    \bar{\bm{c}}_k^{(r)}}_2^2\;.
\end{equation}
The larger the $\text{FDR}^{(r)}$, the more separable clusters
are. Obtaining $\text{FDR}^{(r)}$ is computationally light since
distances in \eqref{eq:samplevar} have been calculated during the
validation phase of the algorithm. The only additional burden is
computing the numerator in \eqref{eq:fFDRours} in
$\mathcal{O}[(\check{d} + \check{d}')K^2]$ complexity. Based on FDR, a
second choice for $f$ is
\begin{equation*}
  f(\mathcal{V}_D^{(r)}) = \bigl\lvert \mathcal{V}_D^{(r)} \bigr\rvert
  \exp\left(-\frac{1}{\text{FDR}^{(r)}}\right)\,.
\end{equation*}
Instead of $\text{FDR}^{(r)}$, the exponent is $-1/\text{FDR}^{(r)}$
to avoid pathological cases where $\text{FDR}^{(r)}$ approaches
$+\infty$, e.g., when all points in a cluster are very concentrated so
that $(\bar{{\sigma}}_k^{(r)})^2 \approx 0$.

Alg.~\ref{alg:SkeVa} incurs overall complexity
$\mathcal{O}(NKR_{\max}\check{d}I)$, where $I$ is an upper bound on
the number of iterations needed for $K$-means to converge in
step~\ref{alg1.step:kmeans}, plus $\mathcal{O}(NKR_{\max}\check{d}')$
required in step~\ref{alg1.step:data.cluster.aug} of
Alg.~\ref{alg:SkeVa}. {Parameters $\check{d}$,
  $\check{d}'$, and $R_{\max}$ are selected depending on the available
  computational resources; $(\check{d}, \check{d}')$ should be such
  that running the computations of Alg.~\ref{alg:SkeVa} on
  $(\check{d}+\check{d}')$-dimensional vectors can be affordable by
  the processing unit used. A probabilistic argument for choosing
  $R_{\max}$ can be determined as elaborated next.}

{
\begin{remark}\label{rem:R.and.D}
  Using parameters that can be obtained in practice, it is possible to
  relate the number of random draws $R_{\max}$ with the reliability of
  SkeVa-based clustering, along the lines of analyzing RANSAC
  \cite{RANSAC81}.

  To this end, let $p$ denote the probability of having out of $R$
  SkeVa realizations \textit{at least one} ``good draw'' of
  $\check{d}$ ``informative'' dimensions, meaning one for which
  $K$-means yields data-cluster associations close to those found by
  $K$-means on the full set of $D$
  dimensions. Parameter $p$ is a function of the
    underlying cluster characteristics. It can be selected by the
  user and reflects one's level of SkeVa-based big data clustering
  reliability, e.g., $p:=0.95$. The probability of having all ``bad
  draws'' after $R$ SkeVa repetitions is clearly $1-p$. Moreover, let
  $q$ denote the probability that a randomly drawn row of $\bm{X}$ is
  ``informative.'' In other words, $q$ quantifies prior information on
  the number of rows (out of $D$) that carry high discriminative
  information for $K$-means. For instance, $q$ can be practically
  defined by the leverage scores of $\bm{X}$, which typically rank the
  importance of rows of $\bm{X}$ in large-scale data analytics
  \cite{Drineas, mahoney2012algorithmic}. An estimate of the leverage
  scores across the rows of $\bm{X}$ expresses $q$ as the percentage
  of informative rows. Alternatively, if $\bm{x}_{n_j} = \bm{m}_k +
  \bm{\Sigma}_k^{1/2}\bm{v}_{n_j}$ is the data generation mechanism
  per cluster $k$, with $\bm{v}_{n_j}\sim \mathcal{N}(\bm{0},
  \bm{I}_D)$, then the $i$th entry of $\bm{x}_{n_j}$ is
  $\bm{e}_i^{\top} \bm{x}_{n_j}\sim \mathcal{N}(\bm{e}_i^{\top}
  \bm{m}_k, [\bm{\Sigma}_k]_{ii})$. Thus, rows of $\bm{X}$ are
  realizations of a Gaussian $1\times N$ random vector. If these rows
  are clustered in $K'$ groups, $q$ can capture the probability of
  having an ``informative'' row located within a confidence region
  around its associated centroid that contains a high percentage
  $\alpha\in(0,1)$ of its pdf mass. Per SkeVa realization, the
  probability of drawing $\check{d}$ ``non-informative'' rows can be
  approximated by $(1-q)^{\check{d}}$. Due to the independence of
  SkeVa realizations, $1-p = (1-q)^{{\check{d}}R}$, which implies that
  $R \simeq \log(1-p) / [\check{d}\log(1-q)]$. Clearly, as
    $R$ increases there is (a growing) nonzero probability that the
    correct clusters will be revealed. On the other hand, it must be
    acknowledged that if $R$ is not sufficient, clustering performance
    will suffer commensurately.

  It is interesting to note that as in \cite{RANSAC81}, $R$ only
  depends implicitly on $D$, since $q$ is a percentage, and does not
  depend on the validation metric or pertinent thresholds and bounds.
\end{remark}
}

\subsection{Sequential algorithm}\label{sec:seq}

\begin{algorithm}
  \begin{algorithmic}[1]
    \algrenewcommand\algorithmicindent{1em}
    \Require{Data $\bm{X}$; number of clusters $K$; reduced dimensions
      $\check{d}$ and $\check{d}'$ of reduced dimensions for the sketching
      and validation phases, respectively; ranking function $f$;
      number of realizations (draws) $R_{\max}$.}
    \Ensure{Data-cluster associations on $\bm{X}$.}
    \For{$r = 1$ to $R_{\max}$}
    \State\parbox[t]{\dimexpr\linewidth-\algorithmicindent}{Randomly sample
      $\check{d} \ll D$ rows of $\bm{X}$ to obtain
      $\check{\bm{X}}^{(r)}$.}\label{alg2.step:sample.dims}
    \State\parbox[t]{\dimexpr\linewidth-\algorithmicindent}{Run $K$-means on
      $\check{\bm{X}}^{(r)}$ to obtain clusters
      $\{\check{\mathcal{C}}_k^{(r)}\}_{k=1}^K$ and centroids
      $\{\check{\bm{c}}_k^{(r)}\}_{k=1}^K$.}\label{alg2.step:kmeans}
    \State\parbox[t]{\dimexpr\linewidth-\algorithmicindent}{Initialize the
      auxiliary set of dimensions $\mathcal{A}=\emptyset$.}
    \For{$j=1$ to $\check{d}'$}
    \State\parbox[t]{\dimexpr\linewidth-2\dimexpr\algorithmicindent}{Randomly sample
      $1$ dimension of $\bm{X}$ (other than those in
      step~\ref{alg2.step:sample.dims} and in $\mathcal{A}$) to obtain row
      $\check{\mathbf{x}}^{(r')}$ of $\bm{X}$.}
    \State\parbox[t]{\dimexpr\linewidth-2\dimexpr\algorithmicindent}{Include
      sampled dimension in $\mathcal{A}$.}
    \State\parbox[t]{\dimexpr\linewidth-2\dimexpr\algorithmicindent}{Form
      $\{\bar{\bm{c}}_k^{(r)}:= [\check{\bm{c}}_k^{(r)}{}^{\top},
      \check{c}_k^{(r')}]^{\top}\}_{k=1}^K$ as in Alg.~\ref{alg:SkeVa}.}
    \State\parbox[t]{\dimexpr\linewidth-2\dimexpr\algorithmicindent}{Associate
      $\{\bar{\bm{x}}_n^{(r)}\}_{n=1}^N$ to closest
      $\{\bar{\bm{c}}_k^{(r)}\}_{k=1}^K$.}\label{alg2.step:data.cluster.aug}
    \State\parbox[t]{\dimexpr\linewidth-2\dimexpr\algorithmicindent}{Identify
      validation set $\mathcal{V}_D^{(r)}$ [cf.\ \eqref{CS}].}
    \If{$f(\mathcal{V}_D^{(r)}) < f_{\max}^{(r)}$ or $|\nabla
      f(\mathcal{V}_D^{(r)})| \le \epsilon$}
    \State\parbox[t]{\dimexpr\linewidth-3\dimexpr\algorithmicindent}{Go to
      step~\ref{alg2.step:sample.dims}.}
    \EndIf
    \EndFor
    \State\parbox[t]{\dimexpr\linewidth-\algorithmicindent}{$f_{\max}^{(r+1)} =
      f(\mathcal{V}_D^{(r)})$.}
    \State\parbox[t]{\dimexpr\linewidth-\algorithmicindent}{$r_* = r$.}
    \EndFor
    \State{Data-cluster associations on $\bm{X}$ according to
      $\{\check{\mathcal{C}}_k^{(r_*)}\}_{k=1}^K$.}
  \end{algorithmic}
  \caption{{Sequential (Se)SkeVa~K-means.}}\label{alg:SkeVaSeq}
\end{algorithm}

{Drawing a batch of $\check{d}'$ features (rows of
  $\bm{X}$) during the validation phase of Alg.~\ref{alg:SkeVa} to
  assess the discriminating ability of the features drawn in the
  sketching phase may be computationally, especially if $\check{d}'$
  is relatively large. The computation of all distances $\{\delta
  (\bar{\bm{x}}_n^{(r)},\bar{\bm{c}}_k^{(r)})\}$ in
  step~\ref{alg1.step:data.cluster.aug} of Alg.~\ref{alg:SkeVa} can be
  prohibitive if $\check{d}'$ becomes large. This motivates a
  sequential augmentation of dimensions, where features are added one
  at a time, and computations are performed on only a single row of
  $\bm{X}$ per feature augmentation, till the upper-bound $\check{d}'$
  is reached. Apparently, such an approach adds flexibility and
  effects computational savings since sequential augmentation of
  dimensions does not need to be carried out till $\check{d}'$ is
  reached, but it may be terminated early on if a prescribed criterion
  is met. These considerations prompted the development of
  Alg.~\ref{alg:SkeVaSeq}.}

The sketching phase of Alg.~\ref{alg:SkeVaSeq} remains the same
as in Alg.~\ref{alg:SkeVa}. In the validation phase, and for each
dimension in the additional $\check{d}'$ ones,
$\{\check{\bm{c}}_k^{(r')}\}_{k=1}^K$ are obtained as in
Alg.~\ref{alg:SkeVa} [cf.\ \eqref{centroid.extra.dims}], and likewise
for $\mathcal{V}_D^{(r)}$. If $f(\mathcal{V}_D^{(r)})$ is smaller than
the current maximum value $f_{\max}^{(r)}$ in memory, the
$\check{d}$-dimensional clustering $\{\check{\mathcal{C}}_k^{(r)}\}$
is discarded, and a new draw is taken. This can be seen as a
``bail-out'' test, to reject possibly ``bad clusterings'' in time,
without having to perform the augmentation using all $\check{d}'$
dimensions.

\begin{figure}
  \centering
  \includegraphics[width=0.7\columnwidth]{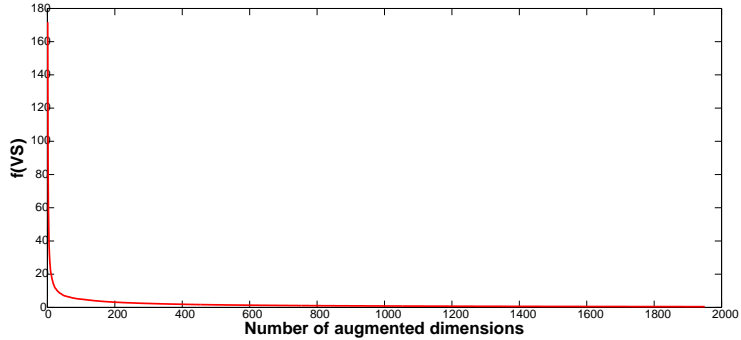}
  \caption{$f(\mathcal{V}_D^{(r)})$ vs.\ the number of augmented
    dimensions $\check{d}'$ for a synthetic data-set, with $D=2,000,
    N=1,000$, $\check{d}=50$, and full-rank data-model (cf.\
    Secs.~\ref{sec:seq} and
    \ref{sec:simulations}).}\label{fig:gradofCS}
\end{figure}

Experiments corroborate that it is not necessary to augment all
$D-\check{d}$ dimensions (cf. Fig.~\ref{fig:gradofCS}), but using a
small subset of them provides satisfactory accuracy while reducing
complexity. An alternative route is to stop augmentation once the
``gradient'' of $f$, meaning finite differences across augmented
dimensions, drops below a prescribed $\epsilon >0$; that is, $|\nabla
f(\mathcal{V}_D^{(r)})|\leq\epsilon$. The sequential approach is
summarized in Alg.~\ref{alg:SkeVaSeq}, and has complexity strictly
smaller than $\mathcal{O}[NKR_{\max}(\check{d}I + \check{d}')]$.

\begin{remark}\label{rem:X.Xadjoint}
  {Using $N$ in the place of $D$ whenever $D\ll N$, or
    equivalently, replacing $\bm{X}\in\Real^{D\times N}$ with
    $\bm{X}^{\top}$, both the batch and sequential schemes developed
    for $D\gg$ can be implemented verbatim for $N\gg$. This variant of
    SkeVa~K-means will be detailed in the next subsection for
    nonlinearly separable clusters.}
\end{remark}

\subsection{Big data kernel clustering}\label{sec:kernel.version}

{A prominent approach to clustering or classifying
  nonlinearly separable data is through \textit{kernels;} see e.g.,
  \cite{bishop}. Vectors $\{\bm{x}_n\}_{n=1}^N$ are mapped to
  $\{\varphi(\bm{x}_n)\}_{n=1}^N$ that live in a higher (possibly
  infinite-) dimensional space $\hilbert$, where inner products
  defining distances in $\hilbert$, using the induced norm
  $\norm{}_{\mathcal{H}} :=
  \innerp{\cdot}{\cdot}_{\mathcal{H}}^{1/2}$, are given by a
  pre-selected (reproducing) kernel function $\kappa$; that is,
  $\innerp{\varphi(\bm{x}_{n})} {\varphi(\bm{x}_{n'})}_{\mathcal{H}} =
  \kappa(\bm{x}_{n}, \bm{x}_{n'})$~\cite{bishop}.  An example of such
  a kernel is the Gaussian one:
  $\kappa_{\bm{\Sigma}}(\bm{x}_n,\bm{x}):= \exp[-(\bm{x} -
  \bm{x}_n)^{\top} \bm{\Sigma}^{-1} (\bm{x} -
  \bm{x}_n)/2]/[(2\pi)^{D/2} (\det\bm{\Sigma})^{1/2}]$.}

For simplicity in exposition, our novel kernel-based (Ke)SkeVa~K-means
approach to big data clustering will be developed for the hard
$K$-means. Extensions to kernel-based soft SkeVa~K-means follow
naturally, and are outlined in
Appendix~\ref{app:soft.kernel.kmeans}. Similar to \eqref{kmeans}, the
kernel-based hard $K$-means proceeds as follows. For
$i\in\{1,2,\ldots,I\}$,
\begin{subequations}\label{kernel.kmeans}
  \begin{align}
    \intertext{[$i$-a] \textbf{Update data-cluster associations:} For
      $n=1,\ldots,N$,} \bm{x}_n\in\mathcal{C}_k[i] & \Leftrightarrow k
    \in \Argmin_{k'\in\{1, \ldots, K\}} \norm*{\varphi(\bm{x}_n) -
      \bm{c}_{k'}[i]}_{\hilbert}^2 \label{kernel.kmeans.a}\\
    \intertext{[$i$-b] \textbf{Update cluster centroids:} For
      $k=1,\ldots,K$,} \bm{c}_k[i+1] & \in \Argmin_{\bm{c}\in\hilbert}
    \sum\nolimits_{\bm{x}_n\in \mathcal{C}_k[i]}
    \norm{\varphi(\bm{x}_n) - \bm{c}}_{\hilbert}^2 \notag\\
    & = \frac{1}{\bigl|\mathcal{C}_k[i]\bigr|}
    \sum\nolimits_{\bm{x}_n\in \mathcal{C}_k[i]} \varphi(\bm{x}_n)
    \label{kernel.kmeans.b}
  \end{align}
\end{subequations}
where Euclidean norms $\norm{}_2$ in the standard form of $K$-means
have been replaced by $\norm{}_{\hilbert}$. As the potentially
infinite-size $\{\bm{c}_k[i+1]\}_{k=1}^K$ cannot be stored in memory,
step \eqref{kernel.kmeans.b} is implicit. In fact, only $\kappa$ and
the data-cluster associations suffice to run \eqref{kernel.kmeans}.
To illustrate this, substitute $\{\bm{c}_k[i+1]\}_{k=1}^K$ from
\eqref{kernel.kmeans.b} into \eqref{kernel.kmeans.a} to write
\begin{align}
  & \norm*{\varphi (\bm{x}_n) -
    \frac{1}{\bigl|\mathcal{C}_{k'}[i+1]\bigr|}
    \sum\nolimits_{\bm{x}'_{n}\in
      \mathcal{C}_{k'}[i+1]} \varphi(\bm{x}'_{n})}_{\hilbert}^2 \notag\\
  & = \innerp{\varphi(\bm{x}_n)}{\varphi(\bm{x}_n)}_{\hilbert} \\
  & \hphantom{=\ } - \frac{2}{\bigl|\mathcal{C}_{k'}[i+1]\bigr|}
  \sum_{\bm{x}'_{n}\in \mathcal{C}_{k'}[i+1]} \innerp{\varphi(\bm{x}_n)}
  {\varphi(\bm{x}'_{n})}_{\hilbert} \notag\\
  & \hphantom{=\ } + \frac{1}{\bigl|\mathcal{C}_{k'}[i+1]\bigr|^2}
  \smashoperator[r]{\sum_{(\bm{x}'_{n}, \bm{x}''_{n})\in
      (\mathcal{C}_{k'}[i+1])^2}} \innerp{\varphi(\bm{x}'_{n})}
  {\varphi(\bm{x}''_{n})}_{\hilbert} \notag\\
  & = \kappa(\bm{x}_n, \bm{x}_n) -
  \frac{2}{\bigl|\mathcal{C}_{k'}[i+1]\bigr|}
  \smashoperator[r]{\sum_{\bm{x}'_{n}\in
      \mathcal{C}_{k'}[i+1]}} \kappa(\bm{x}_n, \bm{x}'_{n}) \notag\\
  & \hphantom{=\ } + \frac{1}{\bigl|\mathcal{C}_{k'}[i+1]\bigr|^2}
  \smashoperator[r]{\sum_{(\bm{x}'_{n}, \bm{x}''_{n})\in
      (\mathcal{C}_{k'}[i+1])^2}} \kappa(\bm{x}'_{n}, \bm{x}''_{n})
  \,. \label{kernel.kmeans.is.affordable}
\end{align}

Having established that distances involved in SkeVa~K-means are
expressible in terms of the chosen kernel $\kappa$, the resulting
iterative scheme is listed as Alg.~\ref{alg:kernel.SkeVa}. After
randomly selecting an affordable subset $\check{\bm{X}}^{(r)}$,
comprising $\check{\nu}$ columns of $\bm{X}$ per realization $r$, and
similar to the trial-and-error step in line~\ref{alg1.step:kmeans} of
Alg.~\ref{alg:SkeVa}, the (kernel) $K$-means of \eqref{kernel.kmeans}
is applied to $\check{\bm{X}}^{(r)}$. The validation phase of
KeSkeVa~K-means is initialized in
line~\ref{alg.kernel.version:start.consensus}, where a second subset
$\check{\bm{X}}^{(r')}$ comprising $\check{\nu}'$ columns from $\bm{X}
\setminus \check{\bm{X}}^{(r)}$. The distances between data
$\{\varphi(\bm{x}_n^{(r')})\}$ and centroids
$\{\check{\bm{c}}_k^{(r)}\}$ involved in
step~\ref{alg.kernel.version:cluster.consensus.2.random.samples} of
Alg.~\ref{alg:kernel.SkeVa} are also obtained through kernel
evaluations [cf.\
\eqref{kernel.kmeans.is.affordable}]. {This KeSkeVa
  that operates on the number of data-points rather than dimensions
  follows along the line of RANSAC \cite{RANSAC81} but with two major
  differences: (i) Instead of robust parameter regression, it is
  tailored for big data clustering; and (ii) rather than consensus it
  deals with affordably small validation sets across possibly huge
  data-sets.}

{During the validation phase, clusters
  $\check{\mathcal{C}}_k^{(r')}$ are specified according to
  $\bm{x}_n^{(r')} \in \check{\mathcal{C}}_k^{(r')} \Leftrightarrow
  k\in\Argmin_{k'\in\{1,\ldots,K\}} \norm{\varphi(\bm{x}_n^{(r')}) -
    \check{\bm{c}}_{k'}^{(r)}}^2_{\hilbert}$. Gathering all
  information from draws $(r,r')$, the augmented clusters
  $\bar{\mathcal{C}}_k^{(r)} := \check{\mathcal{C}}_k^{(r)}
  \cup \check{\mathcal{C}}_k^{(r')}$
  (step~\ref{alg.kernel.version:cluster.consensus.2.random.samples} of
  Alg.~\ref{alg:kernel.SkeVa}) lead to centroids
\begin{equation}
  \bar{\bm{c}}_k^{(r)} := \frac{1}{\bigl\lvert \bar{\mathcal{C}}_k^{(r)}
    \bigr\rvert} \smashoperator[r]{\sum_{\bm{x}_n \in \bar{\mathcal{C}}_k^{(r)}}}
  \varphi(\bm{x}_n)\,. \label{updated.centroids.kernel.version}
\end{equation}
Given the ``implicit centroids'' obtained as in
\eqref{updated.centroids.kernel.version}, data $\bm{X}^{(r)}$ are
mapped to clusters $\{\bar{\mathcal{C}}_k^{(r)}\}$ which are different from
$\check{\mathcal{C}}_k^{(r)}$. To assess this difference, the distance
between $\varphi(\bm{x}_n^{(r)}$ and $\bar{\bm{c}}_k^{(r)}$
is computed, and columns of $\bm{X}^{(r)}$ are re-grouped in clusters
$\{\check{\bar{\mathcal{C}}}_k^{(r)}\}_{k=1}^K$ as
\begin{equation}
  \bm{x}_n^{(r)} \in \check{\bar{\mathcal{C}}}_k^{(r)} \Leftrightarrow k
  \in \Argmin_{k'\in\{1,\ldots,K\}}
  \norm*{\varphi\bigl(\bm{x}_n^{(r)}\bigr) -
    \bar{\bm{c}}_{k'}^{(r)}}^2_{\hilbert}\,. \label{define.kappa.cluster}
\end{equation}
Recall that distances are again obtained through kernel evaluations
[cf.\ \eqref{kernel.kmeans.is.affordable}].}

{The process of generating clusters and centroids in
  KeSkeVa~K-means can be summarized as follows: (i) Group randomly
  drawn data $\bm{X}^{(r)}$ into clusters
  $\check{\mathcal{C}}_k^{(r)}$ with centroids
  $\check{\bm{c}}_k^{(r)}$; (ii) draw additional data-points, augment
  clusters $\bar{\mathcal{C}}_k^{(r)}$, and compute new centroids
  $\bar{\bm{c}}_k^{(r)}$; (iii) given $\bar{\bm{c}}_k^{(r)}$, find
  clusters $\check{\bar{\mathcal{C}}}_k^{(r)}$ as in
  \eqref{define.kappa.cluster}. Since $\check{\bm{c}}_k^{(r)} \neq
  \bar{\bm{c}}_k^{(r)}$ in general, data belonging to
  $\check{\mathcal{C}}_k^{(r)}$ do not necessarily belong to
  $\check{\bar{\mathcal{C}}}_k^{(r)}$, and vice versa; while data
  common to $\check{\bar{\mathcal{C}}}_k^{(r)}$ and
  $\check{\mathcal{C}}_k^{(r)}$, that is data that have not changed
  ``cluster membership'' during the validation phase, comprise the
  \textit{validation set}
\begin{equation}
  \mathcal{V}_N^{(r)}:= \Bigl\{\bm{x}_n^{(r)}\in \check{\bm{X}}^{(r)}
  \Bigm\vert \exists k\ \text{s.t.}\ \bm{x}_n^{(r)}\in
  \Bigl(\check{\mathcal{C}}_k^{(r)} \cap
  \check{\bar{\mathcal{C}}}_k^{(r)}\Bigr)
  \Bigr\}\,. \label{kernel.CS}
\end{equation}
Among $R_{\max}$ realizations, trial $r_*$ with the highest
cardinality $\vert\mathcal{V}_N^{(r_*)}\rvert$ is identified in
Alg.~\ref{alg:kernel.SkeVa}, and data are finally associated with
clusters $\{\check{\mathcal{C}}_k^{(r_*)}\}_{k=1}^K$. The overall
complexity of Alg.~\ref{alg:kernel.SkeVa} is
$\mathcal{O}(DKR_{\max}\check{\nu}^2I)$ in
step~\ref{alg.kernel.version:kmeans}, when $\{\kappa(\bm{x}_n,
\bm{x}_{n'})\}_{n,n'=1}^N$ are not stored in memory and kernel
evaluations have to be performed for all employed data per
realization, plus $\mathcal{O}(DR_{\max}\check{\nu} \check{\nu}')$ in
step~\ref{alg.kernel.version:cluster.consensus.2.random.samples}. If
$\{\kappa(\bm{x}_n, \bm{x}_{n'})\}_{n,n'=1}^N$ are stored in memory,
then Alg.~\ref{alg:kernel.SkeVa} incurs complexity
$\mathcal{O}(KR_{\max}\check{\nu}^2I + R_{\max}\check{\nu}
\check{\nu}')$, which is quadratic only in the small cardinality
$\check{\nu}$.}

\begin{algorithm}
  \begin{algorithmic}[1]
    \algrenewcommand\algorithmicindent{1em}
    \Require{Data $\bm{X}$; number of clusters $K$; number $\check{\nu}$ and
      $\check{\nu}'$ of data during sketching and validation phase,
      respectively; number of realizations $R_{\max}$.}
    \Ensure{Data-cluster associations on $\bm{X}$.}

    \For{$r = 1$ to $R_{\max}$}

    \State\parbox[t]{\dimexpr\linewidth-\algorithmicindent}{Randomly
      sample $\check{\nu}\ll N$ columns of $\bm{X}$ to obtain
      $\check{\bm{X}}^{(r)}$.}\label{alg.kernel.version:rand.sample}

    \State\parbox[t]{\dimexpr\linewidth-\algorithmicindent}{Apply
      $K$-means [cf.\ \eqref{basis.4.kernel.kmeans} and
      \eqref{soft.kmeans}] on
      $\{\varphi(\bm{x}_n^{(r)})\}_{n=1}^{\check{\nu}}$; obtain
      clusters $\{\check{\mathcal{C}}_k^{(r)}\}_{k=1}^K$ and centroids
      $\{\check{\bm{c}}_k^{(r)}\}_{k=1}^K$.}\label{alg.kernel.version:kmeans}

    \State\parbox[t]{\dimexpr\linewidth-\algorithmicindent}{Randomly
      select $\check{\nu}'\ll N$ columns of $\bm{X}$, other than those
      of step~\ref{alg.kernel.version:rand.sample}, to obtain
      $\check{\bm{X}}^{(r')}$.}\label{alg.kernel.version:start.consensus}

    \State\parbox[t]{\dimexpr\linewidth-\algorithmicindent}{Associate
      $\{\varphi(\bm{x}_n^{(r')})\}_{n=1}^{\check{\nu}'}$ to the
      closest centroids $\{\check{\bm{c}}_k^{(r)}\}_{k=1}^K$; obtain
      clusters $\{\bar{\mathcal{C}}_k^{(r)}\}_{k=1}^K$ and centroids
      $\{\bar{\bm{c}}_k^{(r)}\}_{k=1}^K$ [cf.\
      \eqref{updated.centroids.kernel.version}].}
    \label{alg.kernel.version:cluster.consensus.2.random.samples}

    \State\parbox[t]{\dimexpr\linewidth-\algorithmicindent}{Obtain
      clusters $\{\check{\bar{\mathcal{C}}}_k^{(r)}\}$ on
      $\check{\bm{X}}^{(r)}$ according to
      \eqref{define.kappa.cluster}.}

    \State\parbox[t]{\dimexpr\linewidth-\algorithmicindent}{Identify the
      validation set $\mathcal{V}_N^{(r)}$.}

    \EndFor

    \State{$r_*=\arg\min_{r\in\{1,\ldots,R_{\max}\}}
      \lvert\mathcal{V}_N^{(r)}\rvert$.}

    \State\parbox[t]{\dimexpr\linewidth-\algorithmicindent}{Associate
      $\{\varphi(\bm{x}_n)\}_{n=1}^N$ according to the closest centroids
      obtained from clusters $\{\check{\mathcal{C}}_k^{(r_*)}\}_{k=1}^K$.}
  \end{algorithmic}
  \caption{{Kernel (Ke)SkeVa~K-means}}\label{alg:kernel.SkeVa}
\end{algorithm}

One remark is now in order.

\begin{remark}\label{rem:multikernel}
  Similar to all kernel-based approaches, a critical issue is
  selecting the proper kernel -- more a matter of art and prior
  information about the data. Nonetheless, practical so-termed
  \emph{multi-kernel} approaches adopt a dictionary of kernels from
  which one or a few are selected to run $K$-means
  on~\cite{mp2005jmlr}. It will be interesting to investigate whether
  such multi-kernel approaches can be adapted to our SkeVa-based
  operation to alleviate the dependence on a single kernel function.
\end{remark}

\section{Divergence-Based SkeVa}

A limitation of Algs.~\ref{alg:SkeVa} and \ref{alg:SkeVaSeq} is the
trial-and-error strategy which requires clustering per random draw of
samples.
This section introduces
a method to surmount such a need and select a small number of data or
dimensions on which only a \textit{single} clustering step is applied at the
last stage of the algorithm. As the ``quality'' per draw is assessed without
clustering, this approach trades off accuracy for reduced complexity.

\subsection{Large-scale data sets}

First, the case where random draws are performed on the $N$ data will
be examined, followed by draws across the $D$
dimensions. {Any randomly drawn data from
  $\{\bm{x}_n\}_{n=1}^N$ will be assumed centered around their
  sample mean.}

{Since intermediate clustering will not be applied to
  assess the quality of a random draw, a metric is needed to quantify
  how well the randomly drawn samples represent clusters.} To this
end, motivated by the pdf mixture model [cf.~\eqref{log.likelihood}],
consider the following pdf estimate formed using the randomly selected
data
\begin{equation}
  \check{p}^{(r)}(\bm{x}) := \frac{1}{\check{\nu}}
  \sum_{n=1}^{\check{\nu}} \kappa \bigl(\bm{x}_n^{(r)},
  \bm{x}\bigr) \label{def.Parzen.estimate}
\end{equation}
where $\kappa$ stands for a user-defined kernel function,
parameterized by $\bm{x}_n^{(r)}$, and satisfying\\
$\int\kappa(\bm{x}_n^{(r)}, \bm{x})d\bm{x}=1$ for the estimate in
\eqref{def.Parzen.estimate} to qualify as a pdf. Here, the Gaussian
kernel $\kappa_{\bm{\Sigma}}(\bm{x}_n^{(r)},\bm{x}):= \exp[-(\bm{x} -
\bm{x}_n^{(r)})^{\top} \bm{\Sigma}^{-1} (\bm{x} -
\bm{x}_n^{(r)})/2]/[(2\pi)^{D/2} (\det\bm{\Sigma})^{1/2}]$ is
considered with $\bm{\Sigma}:= \sigma^2\bm{I}_D$.

\begin{figure}
  \centering
  \includegraphics[width=0.5\columnwidth]{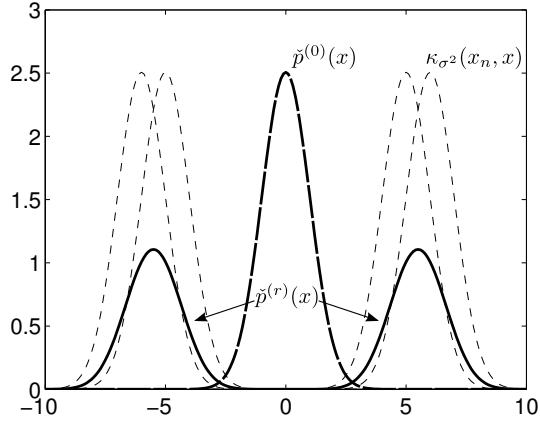}
  \caption{Examples of pdf mixtures fitted to data-points. Reference
    pdf estimate is the uni-modal thick dashed curve; the larger the
    divergence from this estimate, the larger the probability of
    producing meaningful clustering.}\label{fig:multimodal}
\end{figure}

Having linked random samples with pdf estimates, the assessment
whether a draw represents well the whole population
$\{\bm{x}_n\}_{n=1}^N$ translates to how successful this draw is in
estimating the actual data pdf via \eqref{def.Parzen.estimate}. A
random sample where all selected data form a single cluster is clearly
not a good representative of $\{\bm{x}_n\}_{n=1}^N$. For example, if
the selected points are gathered around $\bm{0}$
{(recall that drawn data are centered around their
  sample mean),} then the resulting pdf estimate
  \eqref{def.Parzen.estimate} will resemble the uni-modal (thick)
  dashed curve in Fig.~\ref{fig:multimodal}. Such a pdf estimate is
  a poor representative of the whole data-set, and clustering
  on that small population of data should be avoided. On the contrary,
  random draws yielding the multi-modal (thick) solid curve in
  Fig.~\ref{fig:multimodal} should be highly rated as potential
  candidates for clustering. As a \emph{first step} toward assessing
  draws is a metric quantifying how ``far'' the pdf $\check{p}^{(r)}$
  is from $\check{p}^{(0)}(\bm{x}) := \kappa(\bm{0}, \bm{x})$.

Among candidate metrics of ``distance'' between pdfs, the
Cauchy-Schwarz divergence \cite{Principe.ITL.10} is chosen here:
\begin{equation*}
  \Delta_{\text{CS}} \bigl(\check{p}^{(r)}\distto \check{p}^{(r')}\bigr) := -\log
  \frac{\left(\int \check{p}^{(r)}(\bm{x}) \check{p}^{(r')}(\bm{x})
      d\bm{x} \right)^2}{\int [\check{p}^{(r)}(\bm{x})]^2d\bm{x} \int
    [\check{p}^{(r')}(\bm{x})]^2 d\bm{x}}\,.
\end{equation*}
The reason for choosing $\Delta_{\text{CS}}$ over other popular
divergences, such as the Kullback-Leibler one, is the ease of
obtaining pdf estimates via \eqref{def.Parzen.estimate}. Specifically,
the numerator in $\Delta_{\text{CS}}$ can be expressed as [cf.\
\eqref{def.Parzen.estimate}]
\begin{align}
  \int & \check{p}^{(r)}(\bm{x}) \check{p}^{(r')}(\bm{x}) d\bm{x} \notag\\
  & = \frac{1}{\check{\nu} \check{\nu}'}
  \sum_{n=1}^{\check{\nu}} \sum_{n'=1}^{\check{\nu}'} \int
  \kappa\bigl(\bm{x}_n^{(r)}, \bm{x}\bigr) \kappa\bigl(\bm{x}_{n'}^{(r')},
  \bm{x}\bigr) d\bm{x}\,. \label{numerator.Dcs}
\end{align}
The right-hand-side of \eqref{numerator.Dcs} is simplified further if
the chosen kernel is the Gaussian one. As the convolution of Gaussian
pdfs is also Gaussian, it is not hard to verify that
\begin{equation*}
  \int \kappa_{\bm{\Sigma}} \bigl(\bm{x}_n^{(r)}, \bm{x}\bigr)
  \kappa_{\bm{\Sigma}} \bigl(\bm{x}_{n'}^{(r')}, \bm{x}\bigr) d\bm{x} =
  \kappa_{2\bm{\Sigma}}\bigl(\bm{x}_n^{(r)}, \bm{x}_{n'}^{(r')}\bigr)
\end{equation*}
for which \eqref{numerator.Dcs} becomes
\begin{equation*}
  \int \check{p}^{(r)}(\bm{x}) \check{p}^{(r')}(\bm{x}) d\bm{x} =
  \frac{1}{\check{\nu}\check{\nu}'} \bm{1}^{\top}
  \bm{K}_{2\bm{\Sigma}}^{(r,r')}\bm{1}
\end{equation*}
where the $\check{\nu} \times \check{\nu}'$ matrix
$\bm{K}_{2\bm{\Sigma}}^{(r,r')}$ has $(n,n')$th entry
$[\bm{K}_{2\bm{\Sigma}}^{(r,r')}]_{nn'} :=
\kappa_{2\bm{\Sigma}}(\bm{x}_n^{(r)}, \bm{x}_{n'}^{(r')})$. It thus
follows that
\begin{align}
  \Delta_{\text{CS}} \bigl(\check{p}^{(r)}\distto
  \check{p}^{(r')}\bigr) = & -2\log \left(\frac{1}{\check{\nu}
      \check{\nu}'} \bm{1}^{\top} \bm{K}_{2\bm{\Sigma}}^{(r,
      r')} \bm{1}\right)\notag\\
  & + \log\left(\frac{1}{\check{\nu}^2}\bm{1}^{\top}
    \bm{K}_{2\bm{\Sigma}}^{(r,r)} \bm{1}\right)\notag\\
  & + \log\left(\frac{1}{\check{\nu}'{}^2}\bm{1}^{\top}
    \bm{K}_{2\bm{\Sigma}}^{(r',r')} \bm{1}\right)\,. \label{Dcs.logs}
\end{align}
Notice that when $r'$ is simply $\{\bm{0}\}$, the
last summand in \eqref{Dcs.logs} becomes
$\log\kappa_{2\bm{\Sigma}}(\bm{0},\bm{0}) = -D(\log 2\pi)/2 - (\log\det
2\bm{\Sigma})/2$.

\begin{algorithm}
  \begin{algorithmic}[1]
    \algrenewcommand\algorithmicindent{1em}

    \Require{Data $\bm{X}$; number of clusters $K$; number
      $\check{\nu}$ and $\check{\nu}'$ of points for sketching
      and validation phases, respectively; number of
      realizations $R_{\max}$; $\Delta_{\max}=0$,
      $\Delta'_{\min}=+\infty$.}  \Ensure{Data-cluster associations.}

    \For{$r = 1$ to $R_{\max}$}

    \State\parbox[t]{\dimexpr\linewidth-\algorithmicindent}{Let
      $\check{\bm{X}}^{(r)}$ denote $\check{\nu}$ randomly selected
      points after centered around their sample
      mean.}\label{alg.div.N:rand.sample}

    \If{$\Delta_{\text{CS}}(\check{p}^{(r)}\distto\check{p}^{(0)}) >
      \Delta_{\max}$}\label{alg.div.N:1st.check}

    \State\parbox[t]{\dimexpr\linewidth-3\dimexpr\algorithmicindent}{Let
      $\check{\bm{X}}^{(r')}$ denote $\check{\nu}'$ randomly selected
      points, other than those in step~\ref{alg.div.N:rand.sample},
      after centered around their sample mean.}\label{alg.div.N:begin.aug}

    \State\parbox[t]{\dimexpr\linewidth-3\dimexpr
      \algorithmicindent}{$\bar{\bm{X}}^{(r)}
      := [\check{\bm{X}}^{(r)},
      \check{\bm{X}}^{(r')}]$.}\label{alg.div.N:def.aug.points}

    \If{$\Delta_{\text{CS}}(\bar{p}^{(r)} \distto \check{p}^{(r')}) <
      \Delta'_{\min}$}\label{alg.div.N:2nd.check}

    \State{$\Delta'_{\min} := \Delta_{\text{CS}}(\bar{p}^{(r)} \distto
      \check{p}^{(r')})$.}

    \State{$\Delta_{\max} := \Delta_{\text{CS}}(\check{p}^{(r)}
      \distto\check{p}^{(0)})$.}\label{alg.div.N:def.tau.max}

    \State{$r_* := r$.}

    \EndIf\label{alg.div.N:end.aug}

    \EndIf

    \EndFor

    \State{Perform $K$-means on $\check{\bm{X}}^{(r_*)}$ and associate
      $\bm{X}$ to clusters obtained by $K$-means on
      $\check{\bm{X}}^{(r_*)}$.}
  \end{algorithmic}
  \caption{Divergence (Di)SkeVa~K-means on $N$.}\label{alg:div.N}
\end{algorithm}

The metric in \eqref{Dcs.logs} is computed per draw of the so-termed
divergence-based DiSkeVa~K-means summarized in Alg.~\ref{alg:div.N}. A
number $R_{\max}$ of realizations are attempted to discover a ``good''
draw $r_*$ of data, to which clustering is finally
performed. Line~\ref{alg.div.N:1st.check} in Alg.~\ref{alg:div.N}
checks whether the randomly selected subset yield via
\eqref{def.Parzen.estimate} a pdf $\check{p}^{(r)}$ that differs
enough from the ``singular'' pdf $\check{p}^{(0)} =
\kappa_{\bm{\Sigma}}(\bm{0},\cdot)$. If the divergence exceeds
$\Delta_{\max}$, realization $r$ will be further explored, otherwise
$r+1$ is drawn. Notice that threshold $\Delta_{\max}$ is adaptively
defined and takes, according to line~\ref{alg.div.N:def.tau.max}, the
maximum recorded value from realization $r=0$ till the current one.

If $\check{p}^{(r)}$ passes the first check of being far from
$\check{p}^{(0)}$, lines~\ref{alg.div.N:begin.aug} to
\ref{alg.div.N:end.aug} implement the \emph{second step} of consenting
whether $r$ is indeed a ``good'' realization.  To this end, a number
of $\check{\nu}'$ additional data-points is drawn to form
$\bar{\bm{X}}^{(r)} := [\check{\bm{X}}^{(r)}, \check{\bm{X}}^{(r')}]$
in line~\ref{alg.div.N:def.aug.points}. The mixture pdf
$\bar{p}^{(r)}$ corresponding to $\bar{\bm{X}}^{(r)}$, should stay as
close as possible to $\check{p}^{(r)}$ since reliable pdf estimates
should remain approximately invariant as extra data are added. Drastic
changes of $\Delta_{\text{CS}}$ before and after augmentation suggests
that the draw is likely not to be a good representative of the whole
population. Notice here that $\Delta'_{\min}$ is also adaptively
defined to take the minimum value among all recorded divergences from
the start of iterations. Moreover, both updates of $\Delta'_{\min}$
and $\Delta_{\max}$ are performed once the candidate draw $r$ has
passed through the ``check-points'' of lines~\ref{alg.div.N:1st.check}
and \ref{alg.div.N:2nd.check}.

In the case where the Gaussian kernel is employed,
Alg.~\ref{alg:div.N} has overall complexity
$o[DR_{\max}\check{\nu}^2 +
DR_{\max}(\check{\nu}\check{\nu}' + \check{\nu}'{}^2)]$, if
the kernel matrix $\bm{K}_{2\bm{\Sigma}}$ of all data
$\{\bm{x}_n\}_{n=1}^N$ is not stored in memory and calculations of all
kernel sub-matrices in \eqref{Dcs.logs} are performed per realization;
plus, $\mathcal{O}(D\check{\nu}KI)$ for a single application of
$K$-means on the finally selected draw $r_*$. If
$\bm{K}_{2\bm{\Sigma}}$ is available in memory, then
Alg.~\ref{alg:div.N} incurs complexity
$o[R_{\max}\check{\nu}^2 + R_{\max}(\check{\nu}
\check{\nu}' + \check{\nu}'{}^2) + D\check{\nu}KI]$, that is quadratic
only in the small subset sizes.

{
\begin{remark}\label{rem:R.and.N}
  Along the lines of Remark~\ref{rem:R.and.D}, let $p$ denote the
  probability of having out of $R$ SkeVa realizations at least one
  ``good draw'' of size $\check{\nu}$, meaning one for which $K$-means
  yields centroids close to those found with the ``full data-set.''
  Moreover, let $q$ denote the probability of a datum to lie ``close''
  to its associated centroid. For example, $q$ can capture the
  probability of having a datum located within a confidence region
  which is centered at its associated centroid and contains a high
  percentage of its pdf mass. The probability of having all ``bad
  draws'' is clearly $1-p$. Assuming that data are drawn
  independently, the probability of having one draw contain only data
  located ``far away'' from centroids is $(1-q)^{\check{\nu}}$. Due to
  the independence of random draws in SkeVa, $1-p =
  (1-q)^{{\check{\nu}}R}$, which implies that $R \simeq \log(1-p) /
  [\check{\nu}\log(1-q)]$. Analogous to Remark~\ref{rem:R.and.D}, this
  argument neither involves $N$ nor it depends on the validation
  metric or pertinent thresholds and bounds.
\end{remark}
}

\subsection{High-dimensional data}

Alg.~\ref{alg:div.N} remains operational also when DiSkeVa~K-means deals
with $D\gg$. Although the proposed scheme can be generalized to cope
with both $N\gg$ and $D\gg$, for simplicity of exposition it will be
assumed that only $D\gg$. To this end, consider the following pdf
estimate $\mathbb{R}^{\check{d}}$:
\begin{equation*}
  \check{p}^{(r)}(\check{\bm{x}}) := \frac{1}{N}
  \sum_{n=1}^N \kappa\bigl(\check{\bm{x}}_n^{(r)},
  \check{\bm{x}}\bigr)\,,\quad \forall \check{\bm{x}}\in
  \mathbb{R}^{\check{d}}
\end{equation*}
where $\check{\bm{x}}_n^{(r)}$ denotes a $\check{d}\times 1$ subvector
of the $D\times 1$ vector $\bm{x}_n$.

The counterpart of Alg.~\ref{alg:div.N} on dimensions is listed as
Alg.~\ref{alg:div.D}. Although along the lines of Alg.~\ref{alg:div.N}, there is
a notable difference. In the validation step, where dimensions are increased
(cf.\ line~\ref{alg.div.D:augmentation}) and a pdf estimate is needed for the
augmented set of variables. To define divergence between pdfs of different
dimensions, vectors have to be zero padded from the $\check{d}$-dimensional
$\check{\bm{x}}_n^{(r)}$ to the $(\check{d}+\check{d}')$-dimensional
$\bar{\bm{\chi}}_n^{(r)}:= [\check{\bm{x}}_n^{(r)}{}^{\top},
\bm{0}^{\top}]^{\top}$. Recall here that $\bar{\bm{x}}_n^{(r)} :=
[\check{\bm{x}}_n^{(r)}{}^{\top}, \check{\bm{x}}_n^{(r')}{}^{\top}]^{\top}$. To
avoid confusion, pdf mixtures on these zero-padded vectors are given by
\begin{equation}
  \bar{q}^{(r)}(\bar{\bm{x}}) = \frac{1}{N}
  \sum_{n=1}^N \kappa \bigl(\bar{\bm{\chi}}_n^{(r)},
  \bar{\bm{x}}\bigr)\,,\quad \forall \bar{\bm{x}}\in
  \mathbb{R}^{\check{d}+\check{d}'}\,.\label{def.zero.pad.pdf.mixture}
\end{equation}

Similar to Alg.~\ref{alg:div.N}, the overall complexity of
Alg.~\ref{alg:div.D} is $o[(\check{d} +
\check{d}')N^2R_{\max}]$ for computations in
\eqref{Dcs.logs}, plus $\mathcal{O}(\check{d}NKI)$ for a single
application of $K$-means on the finally selected draw $r_*$.

\begin{algorithm}
  \begin{algorithmic}[1]
    \algrenewcommand\algorithmicindent{1em} \Require{Data $\bm{X}$;
      number of clusters $K$; number $\check{d}$ and $\check{d}'$ of
      dimensions for sketching and validation phases, respectively;
      number of realizations $R_{\max}$; $\Delta_{\max}=0$,
      $\Delta'_{\min}=+\infty$.}  \Ensure{Data-cluster associations.}
    \For{$r = 1$ to $R_{\max}$}
    \State\parbox[t]{\dimexpr\linewidth-\algorithmicindent}{Center
      randomly selected $\check{\bm{X}}^{(r)}$ around their sample
      mean to obtain
      $\check{\bm{X}}^{(r)}$.}\label{alg.div.D:rand.sample}

    \If{$\Delta_{\text{CS}}(\check{p}^{(r)}\distto \check{p}^{(0)}) >
      \Delta_{\max}$}

    \State\parbox[t]{\dimexpr\linewidth-3\dimexpr\algorithmicindent}{Center
      randomly selected $\check{\bm{X}}^{(r')}$, other than those in
      step~\ref{alg.div.D:rand.sample}, around their sample mean to
      obtain $\check{\bm{X}}^{(r')}$.}
    \State\parbox[t]{\dimexpr\linewidth -
      3\dimexpr\algorithmicindent}{Form $\bar{\bm{X}}^{(r)} :=
      [\check{\bm{X}}^{(r)}{}^{\top},
      \check{\bm{X}}^{(r')}{}^{\top}]^{\top}$.}\label{alg.div.D:def.aug.points}
    \State\parbox[t]{\dimexpr\linewidth -
      3\dimexpr\algorithmicindent}{Form $\bar{\bm{\mathcal{X}}}^{(r)}
      := [\check{\bm{X}}^{(r)}{}^{\top}, \bm{0}^{\top}]^{\top}$ and estimate pdf
      mixture $\bar{q}^{(r)}$
      [cf.~\eqref{def.zero.pad.pdf.mixture}].} \label{alg.div.D:augmentation}
    \If{$\Delta_{\text{CS}}(\bar{p}^{(r)} \distto \bar{q}^{(r)}) < \Delta'_{\min}$}
    \State{$\Delta'_{\min} := \Delta_{\text{CS}}(\bar{p}^{(r)} \distto \bar{q}^{(r)})$.}
    \State{$\Delta_{\max} := \Delta_{\text{CS}}(\check{p}^{(r)}\distto \check{p}^{(0)})$.}
    \State{$r_* := r$.}
    \EndIf

    \EndIf
    \EndFor
    \State{Perform $K$-means on $\check{\bm{X}}^{(r_*)}$ and associate
      $\bm{X}$ to clusters obtained by $K$-means on
      $\check{\bm{X}}^{(r_*)}$.}
  \end{algorithmic}
  \caption{Divergence (Di)SkeVa~K-means on $D$.}\label{alg:div.D}
\end{algorithm}

\begin{remark}\label{rem:mapreduce}
  If multi-core machines are also available, the validation phases of
  Algs.~\ref{alg:SkeVa}-\ref{alg:div.D} can be readily parallelized,
  using recent advances on efficient parallel computing platforms such
  as MapReduce~\cite{dean2008mapreduce, elgohary.icdm.14}.
\end{remark}

{
\section{Numerical Tests}\label{sec:simulations}

We validated the proposed algorithms on synthetic and real
data-sets. Tests involve either large number of data ($N \gg$) and/or
large number of dimensions ($D\gg$). The following methods were also
tested: (i) The standard hard $K$-means [cf.~\eqref{kmeans}], run on
the full range of $N$ data-points and $D$ dimensions, which is
abbreviated in the figures as ``full $K$-means''; (ii) the
state-of-the-art RP-based \textit{feature-extraction} scheme
\cite[Alg.~2]{Drineas}, with a Bernoulli-type RP matrix as in
\cite{achlioptas2001database}; (iii) the \textit{randomized
  feature-selection} (RFS) algorithm \cite[Alg.~1; $\epsilon =
1/3$]{Drineas}, a leverage-scores-based scheme; and (iv) the
\textit{``approximate kernel $K$-means''} algorithm
\cite[Alg.~2]{chitta.kdd.11}, which solves for an ``optimal''
data-cluster association matrix given a randomly selected subset of
the original data-points. For fairness, the naive kernel $K$-means
algorithm in \cite[Alg.~1]{chitta.kdd.11} is not tested, because a
random draw of data and the application of $K$-means is done only
\textit{once} in \cite[Alg.~1]{chitta.kdd.11}; hence, the attractive
attribute of multiple independent draws is not leveraged as in
SkeVa. To mitigate initialization-dependent performance, each
realization of $K$-means, including also its usage as a module in
other competing methods, is run five times with different
initialization per run, keeping finally only the data-clusters
association that results with the smallest sum of distances of data
from the associated centroids.

As figures of merit we adopted the relative clustering accuracy and
the execution time (in secs). Relative clustering accuracy is defined
as the percentage of points assigned to the correct clusters
(empirical probability of correct clustering), relative to that of
(kernel) $K$-means on the full data-set. Regarding computational time
evaluations, tests in Sec.~\ref{sec:simdim} are run using
Matlab~\cite{MATLAB:2013} on a SunFire X$4600$ PC with a $32$-core AMD
Opteron $8356$, clocked at $2.3$GHz with $128$GB RAM
memory~\cite{MSI}, without the use of parallelization, on a single
computational thread. Tests in Secs.~\ref{sec:simpoints},
\ref{sec:simkernel} and \ref{sec:multithread} are run on an HP
ProLiant BL$280$c G$6$ server using $2$ eight-core Sandy Bridge
E$5$-$2670$ processor chips ($2.6$GHz) and $128$GB of RAM
memory~\cite{MSI}. In the latter tests, algorithms were allowed to
exploit MATLAB's inherent multithread
capabilities~\cite{matlabmultithread} on the $16$ cores of the
server. Moreover, all plotted curves are averages over $50$
Monte~Carlo realizations.

To construct synthetic data, $D\times 1$ vectors
$\{\bm{x}_n\}_{n=1}^N$ were generated according to the following model
per cluster $k$:
\begin{equation}
  \bm{x}_{n_j} = \bm{m}_k + \bm{\Sigma}_k^{1/2}\bm{v}_{n_j}\,, \quad
  j\in \{1, \ldots, {N}/{K}\} \label{model.synth.data}
\end{equation}
where it is assumed that $N$ is an integer multiple of $K$, $\bm{m}_k$
is the $D\times 1$ mean (centroid) of cluster $k$, noise
$\bm{v}_{n_j}\sim\mathcal{N}(\bm{0}, \bm{I}_D)$ is standardized
Gaussian, and $\bm{\Sigma}_k$ is the covariance matrix of the data
generated for cluster $k$; hence, $\bm{x}_{n_j} \sim
\mathcal{N}(\bm{m}_k, \bm{\Sigma}_k)$. Means $\{\bm{m}_k\}_{k=1}^K$
are selected uniformly at random from a $D$-dimensional hypercube, as
in \cite{Drineas}. To accommodate data-models with limited degrees of
freedom, the ``rank of data,'' controlled by the number of non-zero
eigenvalues of $\bm{\Sigma}_k$, was used as a tuning parameter. In
certain cases, clusters were well separated---a scenario where
$K$-means achieves relatively high clustering accuracy. Throughout
this section $R_{\max}=10$ except for the tests using the KDDb
database \cite{kddb2010} for which $R_{\max} = 20$.

\subsection{Large number of dimensions $(D\gg)$}\label{sec:simdim}

Tests cases in this subsection have $D\gg N$. Competing methods are
the ``full $K$-means,'' RP \cite[Alg.~2]{Drineas}, and RFS
\cite[Alg.~1; $\epsilon = 1/3$]{Drineas}. Model
\eqref{model.synth.data} was used to generate $N = 1,000$
$D$-dimensional vectors for $K=5$ clusters, for several values of $D$,
and variable ``data-rank.'' It can be seen from
Figs.~\ref{fig:normalandseqfullrank} and \ref{fig:normalandseqlowrank}
that Algs.~\ref{alg:SkeVa} (SkeVa~K-means) and \ref{alg:SkeVaSeq}
(SeSkeVa~K-means) approach the accuracy of the full $K$-means
algorithm as the number of sampled dimensions $d$ increases. As
expected, computational time is significantly lower than that of full
$K$-means, since the latter operates on all $D$ dimensions. Moreover,
SeSkeVa~K-means needs more time than RP~\cite{Drineas} to achieve the
same clustering accuracy (cf.\ Figs.~\ref{fig:normalandseqfullrank}
and \ref{fig:normalandseqlowrank}), since RP utilizes $K$-means as a
sub-module only once, after dimensionality-reduction has been effected
by left-multiplication of $\bm{X}$ with a (fat) $d\times D$ RP
matrix. However, this changes as $D$ grows
large. As Fig.~\ref{fig:hugeD} demonstrates, whenever $D$ is massive,
left-multiplying $\bm{X}$ by the RP $d\times D$ matrix in
\cite{Drineas} can become cumbersome, resulting in computational times
larger than those of SeSkeVa~K-means.

\begin{figure}
  \centering
  \subfloat[Relative clustering accuracy]
  {\includegraphics[width=0.5\columnwidth]{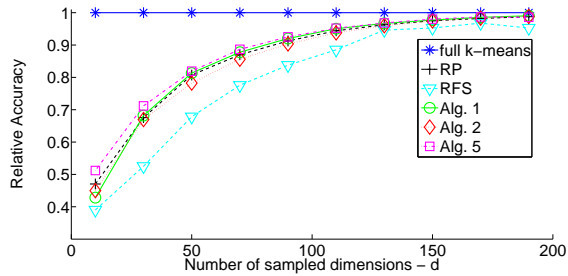}%
    \label{fig:normalandseqfullrank_acc}}\\
  \centering
  \subfloat[Clustering time (secs)]
  {\includegraphics[width=0.5\columnwidth]{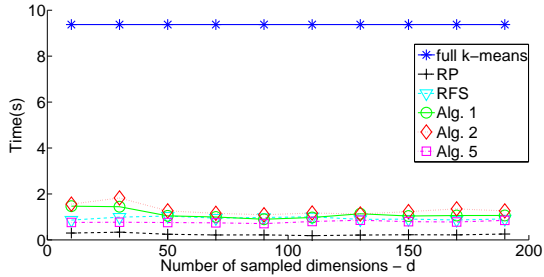}%
    \label{fig:normalandseqfullrank_time}}
  \caption{Synthetic data ($D=2,000$ and full-rank
    model).}\label{fig:normalandseqfullrank}
\end{figure}%

\begin{figure}
  \centering
  \subfloat[Relative clustering accuracy]
  {\includegraphics[width=0.5\columnwidth]{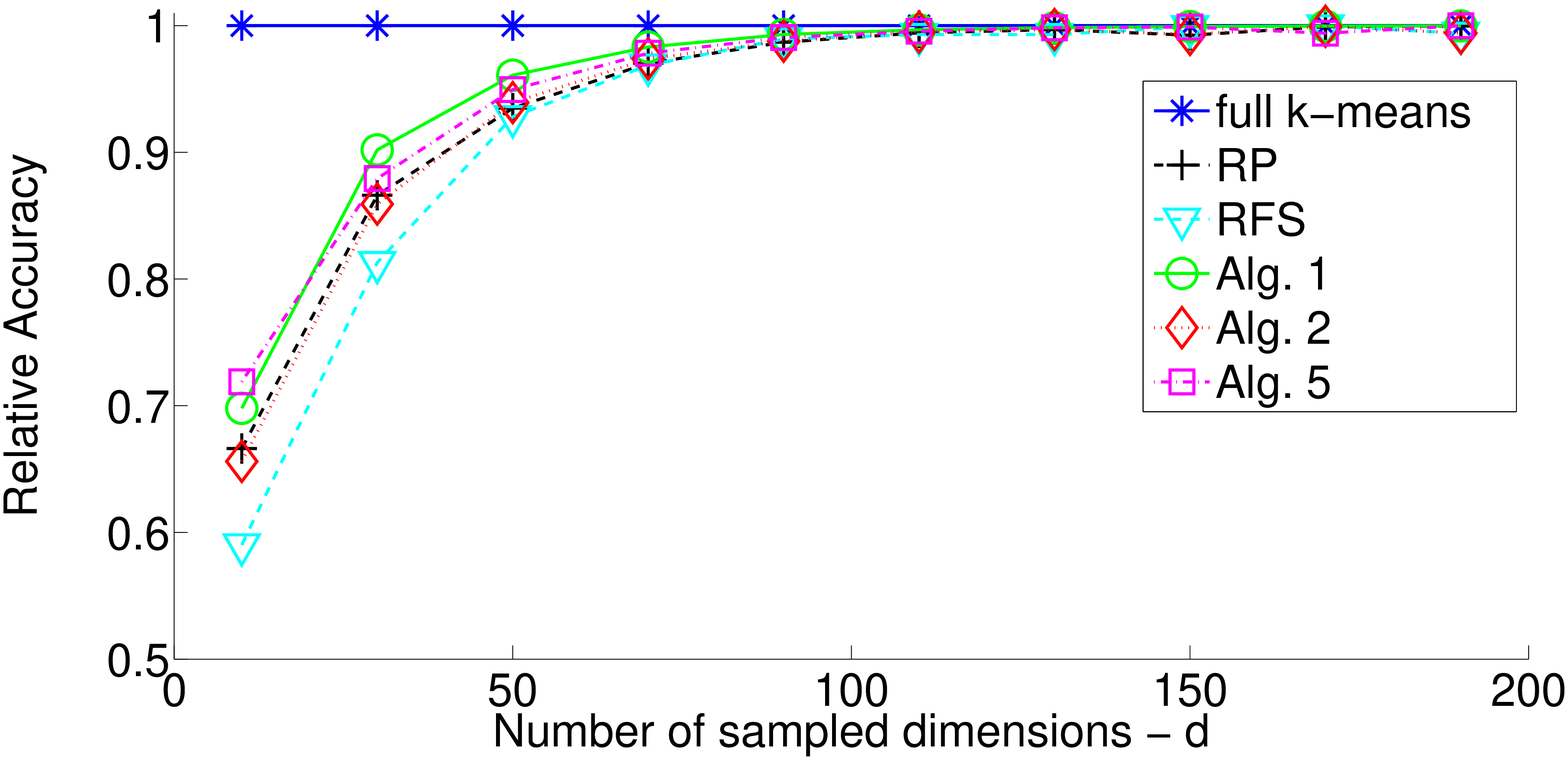}%
    \label{fig:normalandseqlowrank_acc}}\\
  \centering
  \subfloat[Clustering time (secs)]
  {\includegraphics[width=0.5\columnwidth]{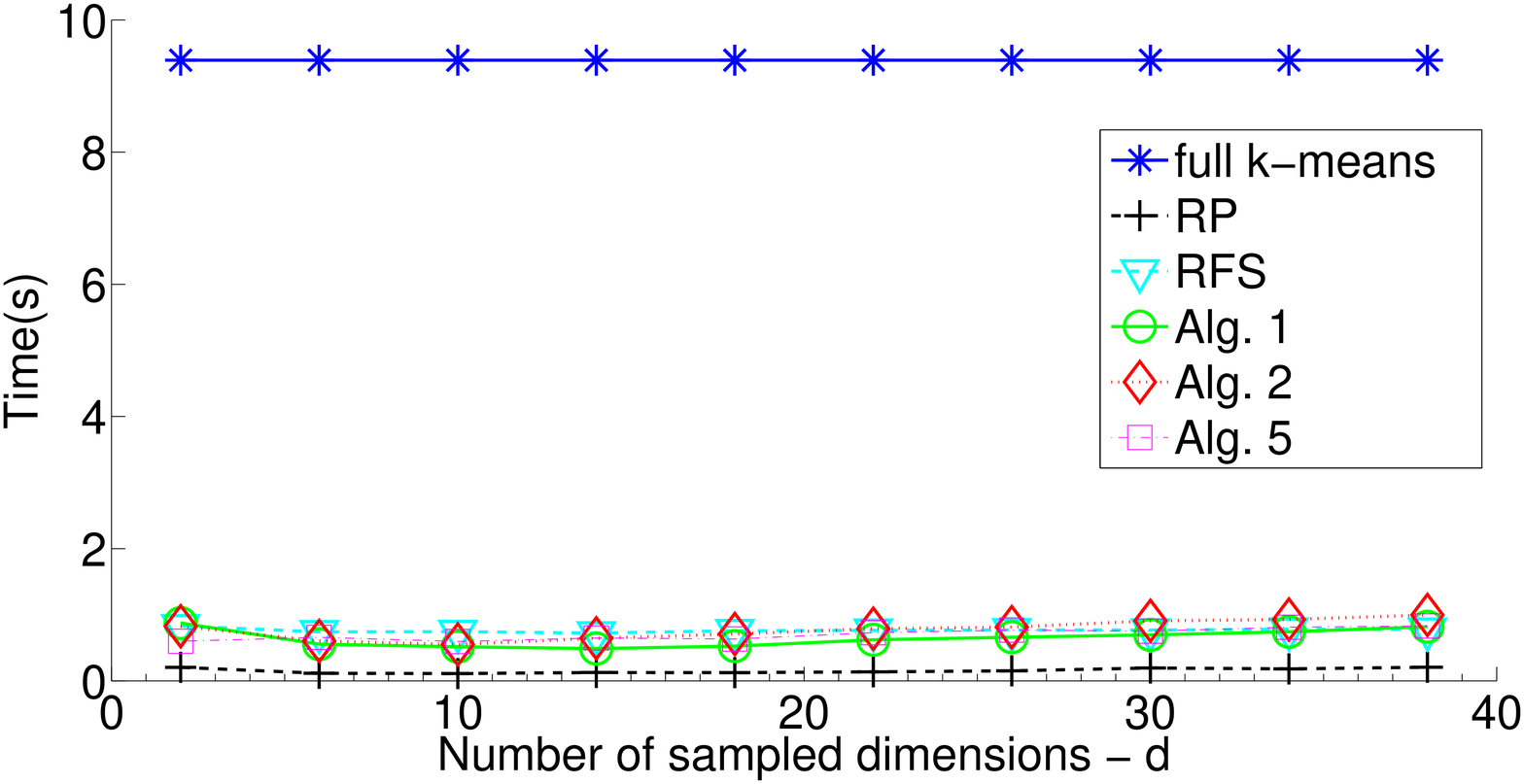}%
    \label{fig:normalandseqlowrank_time}}
  \caption{Synthetic data ($D=2,000$ and rank equal to
    $500$).}\label{fig:normalandseqlowrank}
\end{figure}

\begin{figure}
  \centering
  \subfloat[Relative clustering accuracy]
  {\includegraphics[width=0.5\columnwidth]{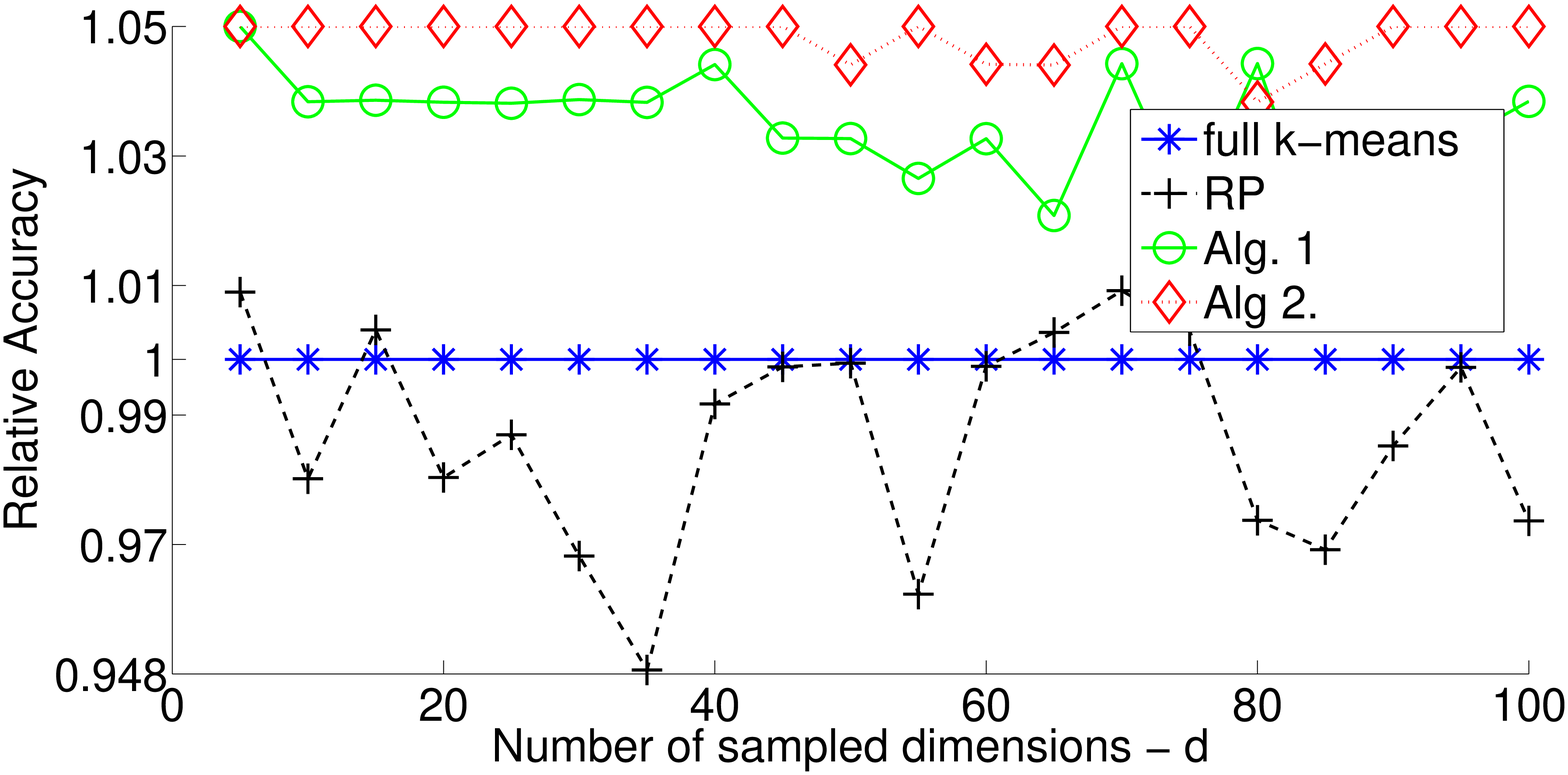}%
    \label{fig:hugeD_acc}}\\
  \centering
  \subfloat[Clustering time (secs)]
  {\includegraphics[width=0.5\columnwidth]{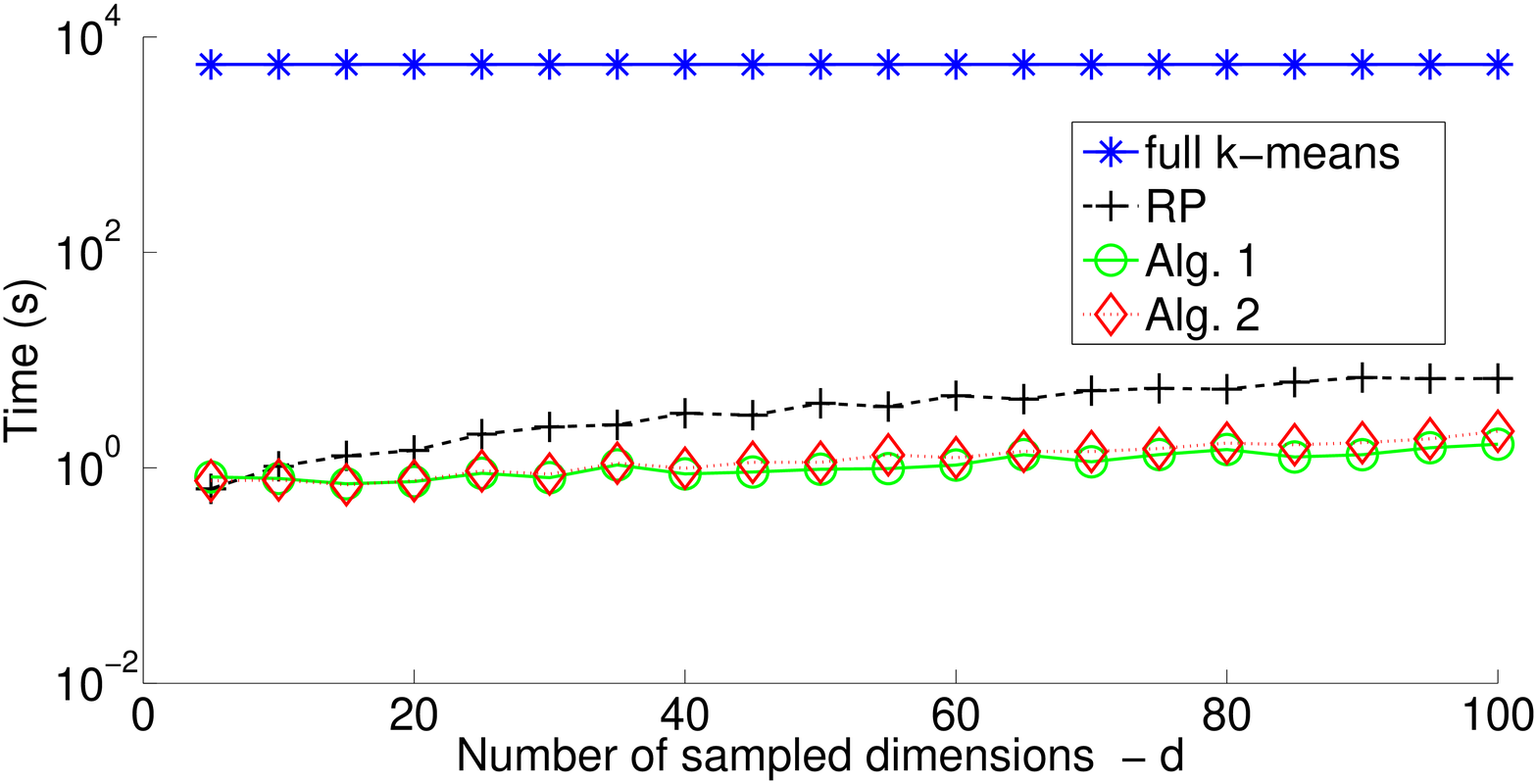}%
    \label{fig:hugeD_time}}
  \caption{Synthetic data ($D=500,000$ and rank equal to
    $1,000$).}\label{fig:hugeD}
\end{figure}

Fig.~\ref{fig:arcene} shows results for the real data-set
ARCENE~\cite{ARCENEguyon2004result}, which contains mass-spectra of
patients diagnosed with cancer, as well as spectra associated with
healthy individuals. Clustering involves grouping $100$
($D=10,000$)-dimensional spectra in two clusters ($K=2$). The number
of augmented dimensions for all employed algorithms is
$\check{d}'=100$. All proposed algorithms approach the performance of
RP and full $K$-means, while requiring less time. Alg.~\ref{alg:div.D}
is the fastest one at a comparable performance.

Fig.~\ref{fig:orl} depicts results for the real ORL database
of $400$ face-images, from $40$ different subjects ($10$ each)
\cite{ORL.paper.94, ORLdb}. Images have size $92\times 112$ with
$8$-bit grey levels, resulting in $D=10,304$. Only $30$ images ($3$
subjects) were used, and as such the task is to group these images
into $K=3$ clusters. As with the ARCENE data, the number of additional
dimensions for all proposed algorithms is
$\check{d}'=100$. Algs.~\ref{alg:SkeVa}, \ref{alg:SkeVaSeq}, and
\ref{alg:div.D} exhibit similar performance, while requiring much less
time than the full $K$-means. Again, Alg.~\ref{alg:div.D} is the
fastest one at a comparable performance.

Tests were also performed on a subset of the KDDb 2010 data-set ($K=2$,
$D=2,990,384$)~\cite{kddb2010}. The version of the data-set is the one
transformed by the winner of the KDD 2010 Cup (National Taiwan
University). In each run $10,000$ data-points were chosen randomly from
both classes, and clustering was performed on this subset. The RFS
performance is not reported in Fig.~\ref{fig:kddb_seq} as there were
issues regarding memory usage due to the required SVD
computations. Here, the number of augmented dimensions is
$\check{d}'=1,000$.  All algorithms exhibit performance similar to
full $K$-means; however Alg.~\ref{alg:SkeVa} and Alg.~\ref{alg:div.D}
require significantly less time than all competing alternatives.

It should be noted also that the required amount of memory per
iteration for Algs.~\ref{alg:SkeVa} and \ref{alg:SkeVaSeq} is at most
$\mathcal{O}[N(\check{d} + \check{d}')]$, in contrast with the
competing algorithms whose memory requirements grow linearly with $D$.

\begin{figure}
  \centering
  \subfloat[Relative clustering accuracy]
  {\includegraphics[width=0.5\columnwidth]{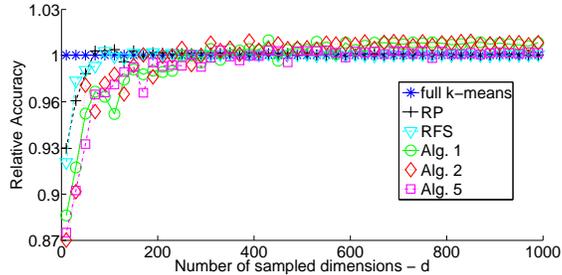}
    \label{fig:arceneacc}}\\
  \centering
  \subfloat[Clustering time (secs)]
  {\includegraphics[width=0.5\columnwidth]{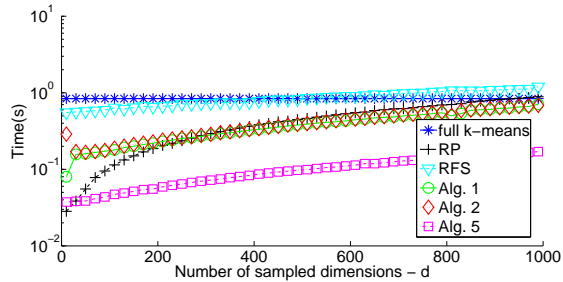}
    \label{fig:arcenetime}}
  \caption{Simulated performance for real data-set ARCENE.}
  \label{fig:arcene}
\end{figure}

\begin{figure}
  \centering
  \subfloat[Relative clustering accuracy]
  {\includegraphics[width=0.5\columnwidth]{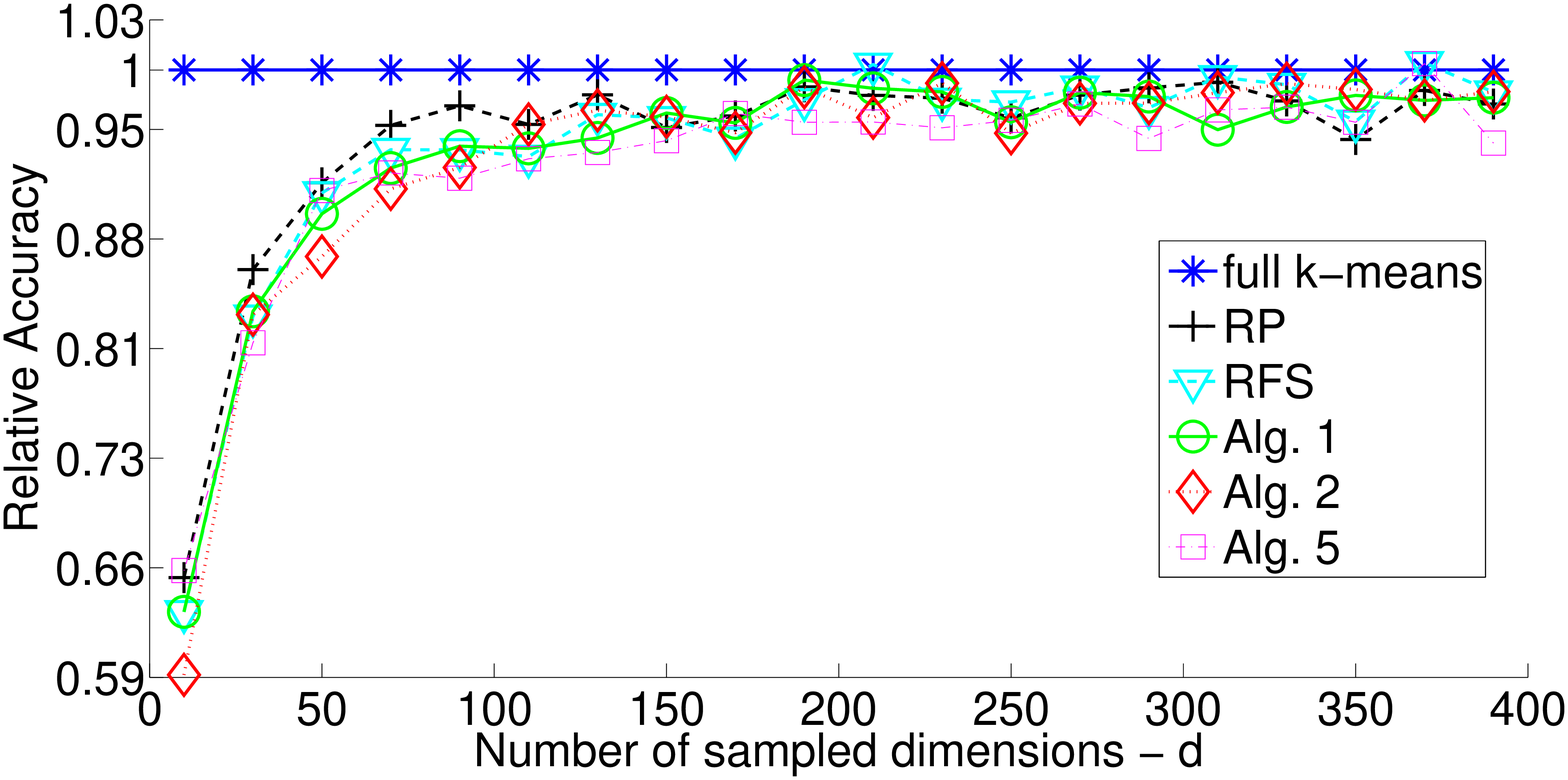}\label{fig:orlacc}}\\
  \centering
  \subfloat[Clustering time (secs)]
  {\includegraphics[width=0.5\columnwidth]{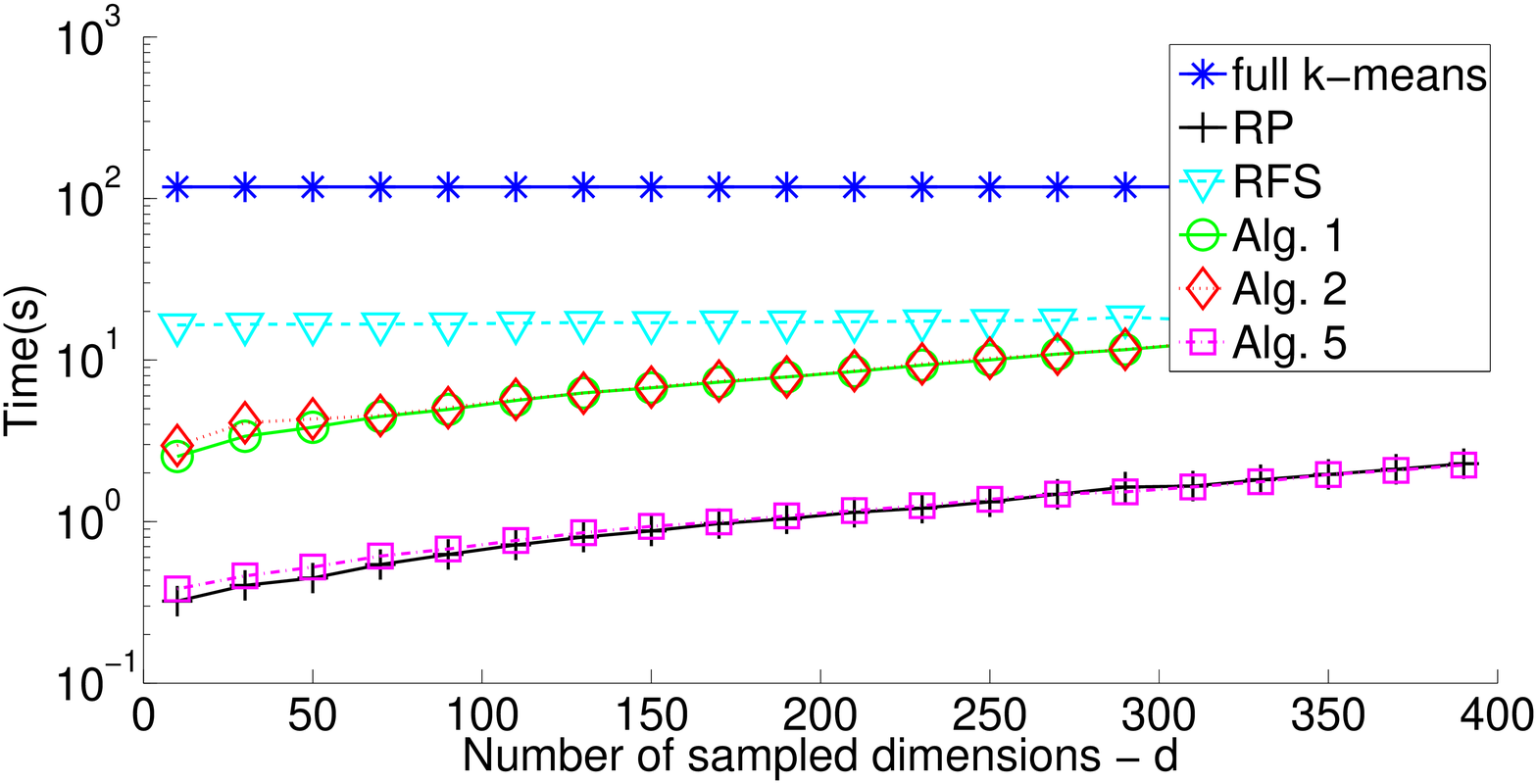}\label{fig:orltime}}
  \caption{Simulated performance for real data-set ORL.}
  \label{fig:orl}
\end{figure}

\begin{figure}
  \centering
  \subfloat[Relative clustering accuracy]
  {\includegraphics[width=0.5\columnwidth]{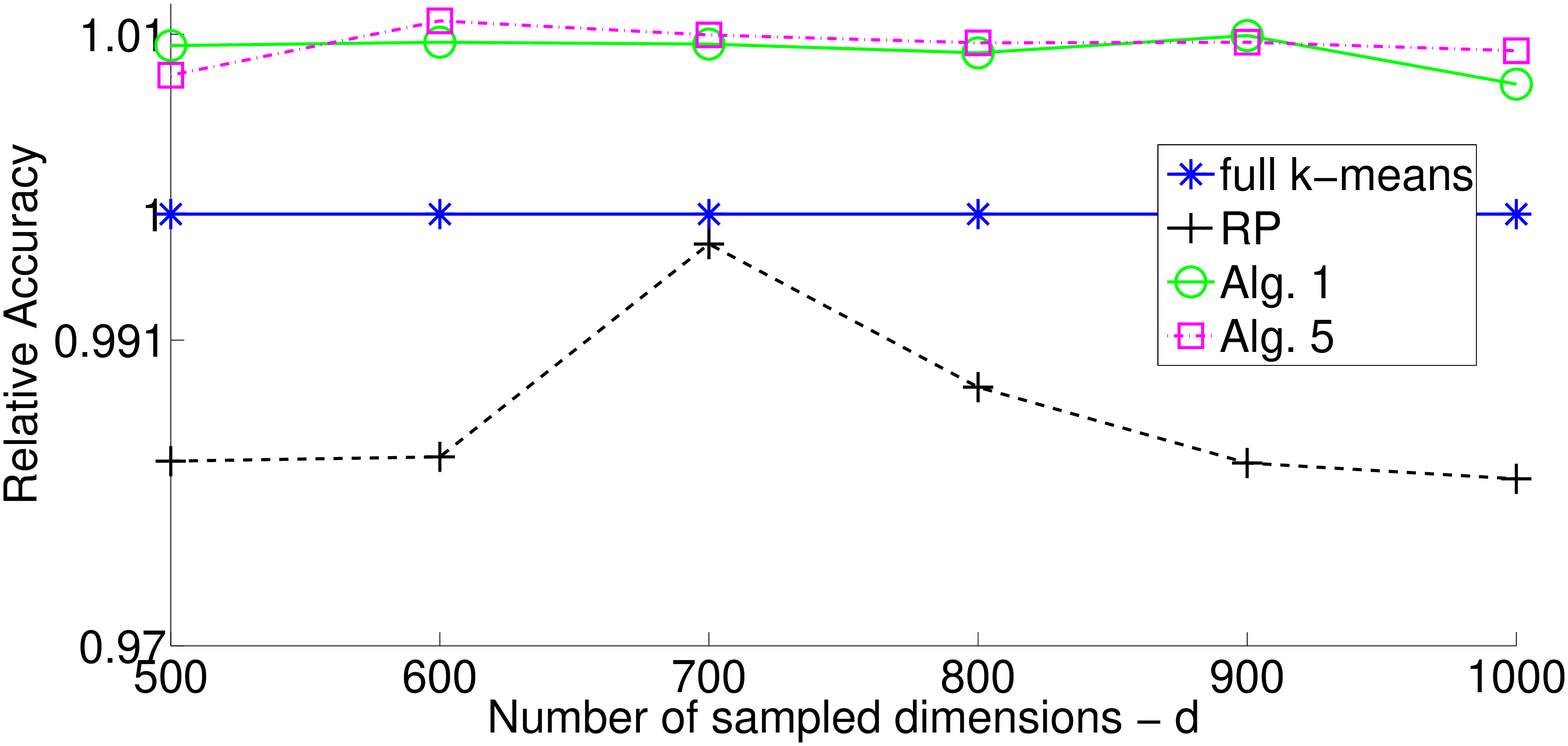}
\label{fig:kddb_seqacc}}\\
  \centering
  \subfloat[Clustering time (secs)]
  {\includegraphics[width=0.5\columnwidth]{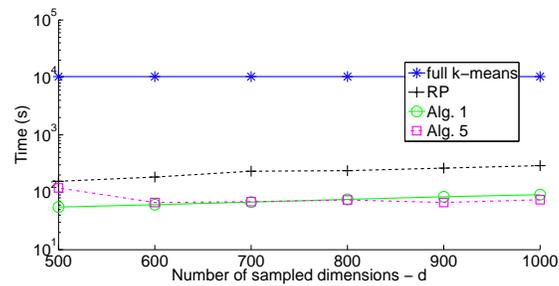}
\label{fig:kddb_seqtime}}
  \caption{Simulated performance for real data-set KDDb.}
  \label{fig:kddb_seq}
\end{figure}

\subsection{Large number of points $(N\gg)$}\label{sec:simpoints}

Here we deal with $N\gg D$. Alg.~\ref{alg:div.N} is compared with the
``full $K$-means'' and the ``approximate kernel $K$-means''
algorithm~\cite{chitta.kdd.11}. $N=100,000$ vectors with $D=5$ were
generated according to \eqref{model.synth.data} for $K=5$
clusters. Although ``approximate kernel $K$-means'' can accommodate
nonlinearly separable clusters by using nonlinear kernel
functions, the linear kernel was used here: $\kappa(\bm{x}, \bm{y}) :=
\bm{x}^{\top} \bm{y}$, $\forall \bm{x}, \bm{y}$. The Gaussian kernel
function $\kappa_{\bm{\Sigma}}$, with $\bm{\Sigma}:= \bm{I}_D$, was
used in \eqref{def.Parzen.estimate}. Fig.~\ref{fig:points100000D5}
shows clustering accuracy across the number $\check{\nu}$ of randomly
selected data per draw. The number of additional points $\check{\nu}'$
is set equal to $100$. As Fig.~\ref{fig:points100000D5} demonstrates,
Alg.~\ref{alg:div.N} approaches the performance of the full $K$-means
algorithm, even with $\check{\nu}=100$ sampled data, while requiring
markedly lower execution time than both ``full'' and ``approximate
kernel $K$-means.''

\subsection{Kernel clustering}\label{sec:simkernel}

Nonlinearly separable data are mapped here using a prescribed kernel
function to high-dimensional spaces to render them linearly
separable. Algs.~\ref{alg:kernel.SkeVa} and \ref{alg:div.N}, with
kernel $K$-means applied only at the end, are compared with the ``full
kernel $K$-means'' and the ``approximate kernel
$K$-means''~\cite{chitta.kdd.11}. Throughout, $\bm{\Sigma} :=
5\bm{I}_D$ was used in \eqref{def.Parzen.estimate}. Tests were
performed on a subset ($N=35,000$) of the MNIST-$784$ data-set, which
contains $28\times 28$ pixel images of handwritten digits grouped in
$K=10$ clusters. The kernel used in this case is the sigmoid one
$\kappa(\bm{x},\bm{y}) = \tanh(\alpha\bm{x}^{\top}\bm{y} + b)$ with
parameters $\alpha = 0.0045$, $b=0.11$, in accordance
to~\cite{zhang2002large}. Both the sigmoid and the Gaussian (for the
case of Alg.~\ref{alg:div.N}) kernels are considered stored in
memory. Fig.~\ref{fig:knkmeansmnist} depicts the relative clustering
accuracy for this data-set and the required time in seconds. It is clear
that accuracies of all three algorithms are close and
approach the performance of ``full kernel $K$-means'' as the number
of sampled data-points increases. However, the time required by
Algs.~\ref{alg:kernel.SkeVa} and \ref{alg:div.N} is significantly less
than the time required by the ``full'' and ``approximate kernel
$K$-means.''

\subsection{Exploiting multiple computational
  threads} \label{sec:multithread}

To showcase the scalability of the proposed algorithms in the presence
of multiple computational nodes, the algorithms were run on multiple
computational threads. The independent draws $r$ of the proposed
algorithms were executed in parallel. Moreover, competing algorithms
were allowed to exploit MATLAB's multithread capabilities, e.g.,
matrix-matrix multiplications in
RP~\cite{matlabmultithread}. Figs.~\ref{fig:arcene_par} and
\ref{fig:kddb_par} show simulation results for the ARCENE and KDDb
data-sets, respectively. Clearly, parallelization of the iterations on
the proposed algorithms is beneficial since the algorithms exhibit
about an order of magnitude less required time than that of competing
methods.

\begin{figure}
  \centering
  \subfloat[Relative clustering accuracy]
  {\includegraphics[width=0.5\columnwidth]{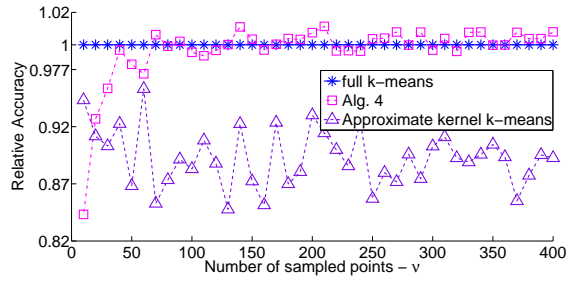}
    \label{fig:points100000D5acc}}\\
  \centering
  \subfloat[Clustering time (secs)]
  {\includegraphics[width=0.5\columnwidth]{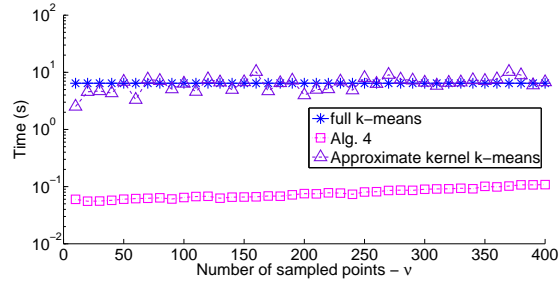}
    \label{fig:points100000D5time}}
  \caption{Synthetic data ($D=5$, $N=100,000$, $K=5$).}
  \label{fig:points100000D5}
\end{figure}



\begin{figure}
  \centering
  \subfloat[Relative clustering accuracy]
  {\includegraphics[width=0.5\columnwidth]{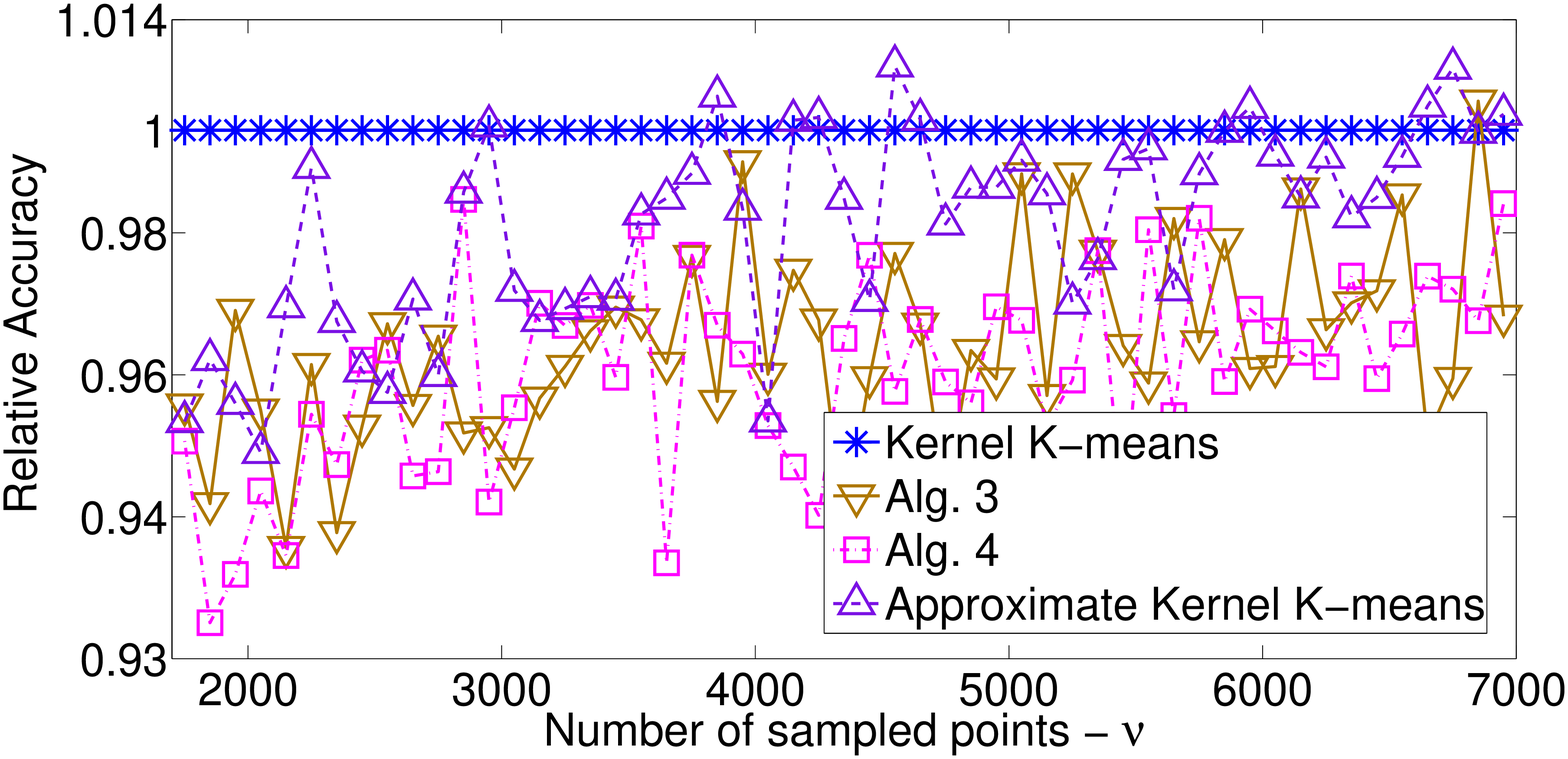}
    \label{fig:knkmeansmnistacc}}\\
  \centering
  \subfloat[Clustering time (secs)]
  {\includegraphics[width=0.5\columnwidth]{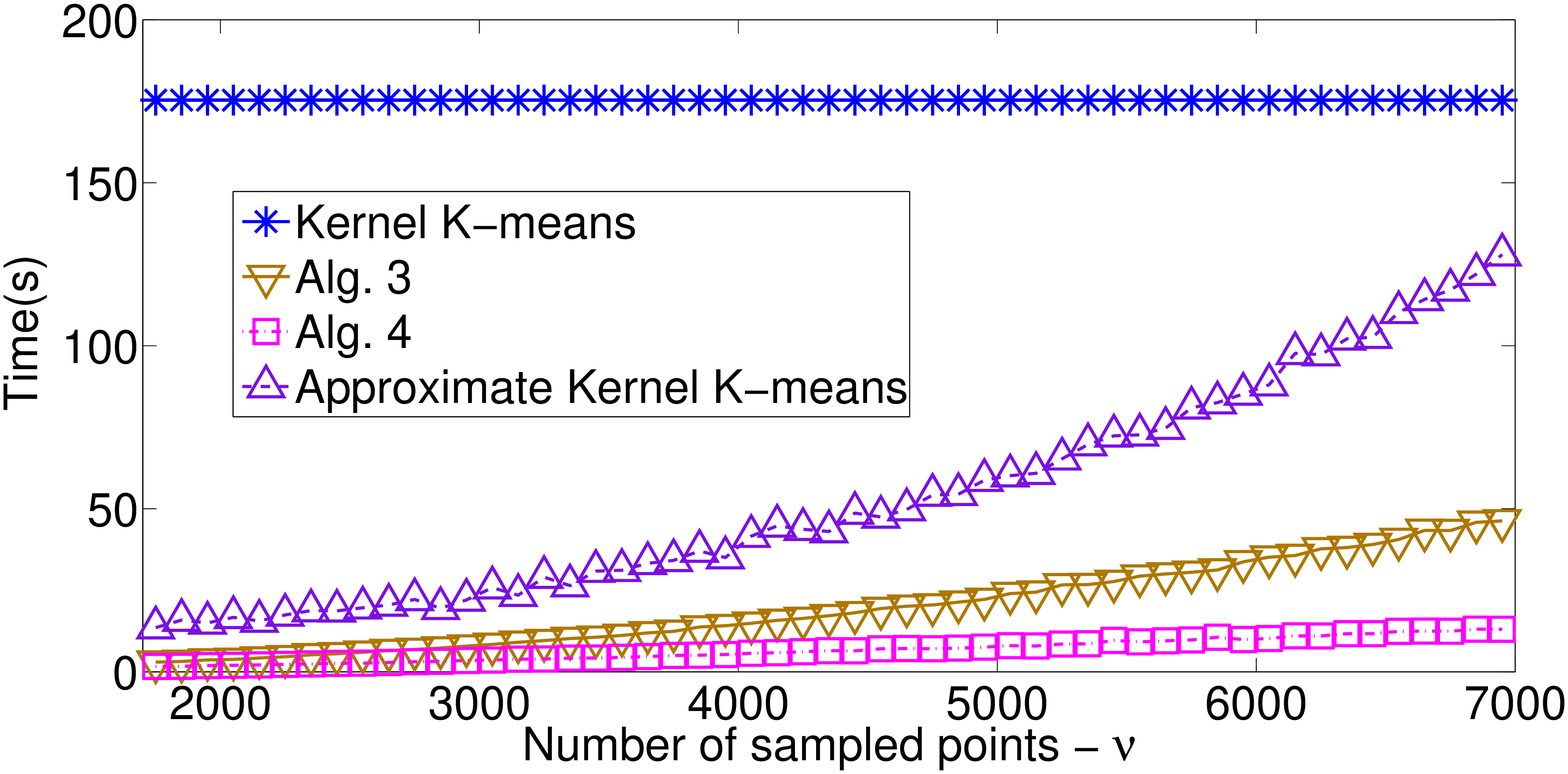}
    \label{fig:knkmeansmnisttime}}
  \caption{A subset of the MNIST data-set.}
  \label{fig:knkmeansmnist}
\end{figure}

\begin{figure}
  \centering
  \subfloat[Relative clustering accuracy]
  {\includegraphics[width=0.5\columnwidth]{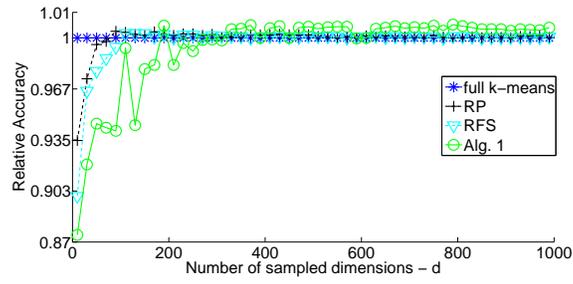}
    \label{fig:arcene_par_acc}}\\
  \centering
  \subfloat[Clustering time (secs)]
  {\includegraphics[width=0.5\columnwidth]{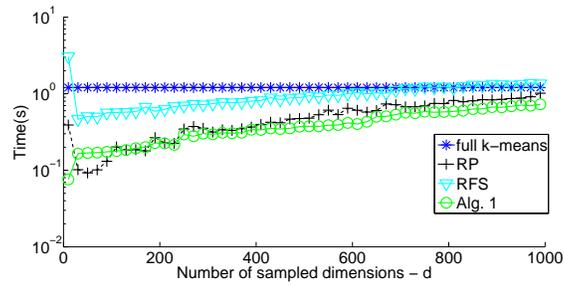}
    \label{fig:arcene_par_time}}
  \caption{A subset of the ARCENE data-set with multithreading.}
  \label{fig:arcene_par}
\end{figure}

\begin{figure}
  \centering
  \subfloat[Relative clustering accuracy]
  {\includegraphics[width=0.5\columnwidth]{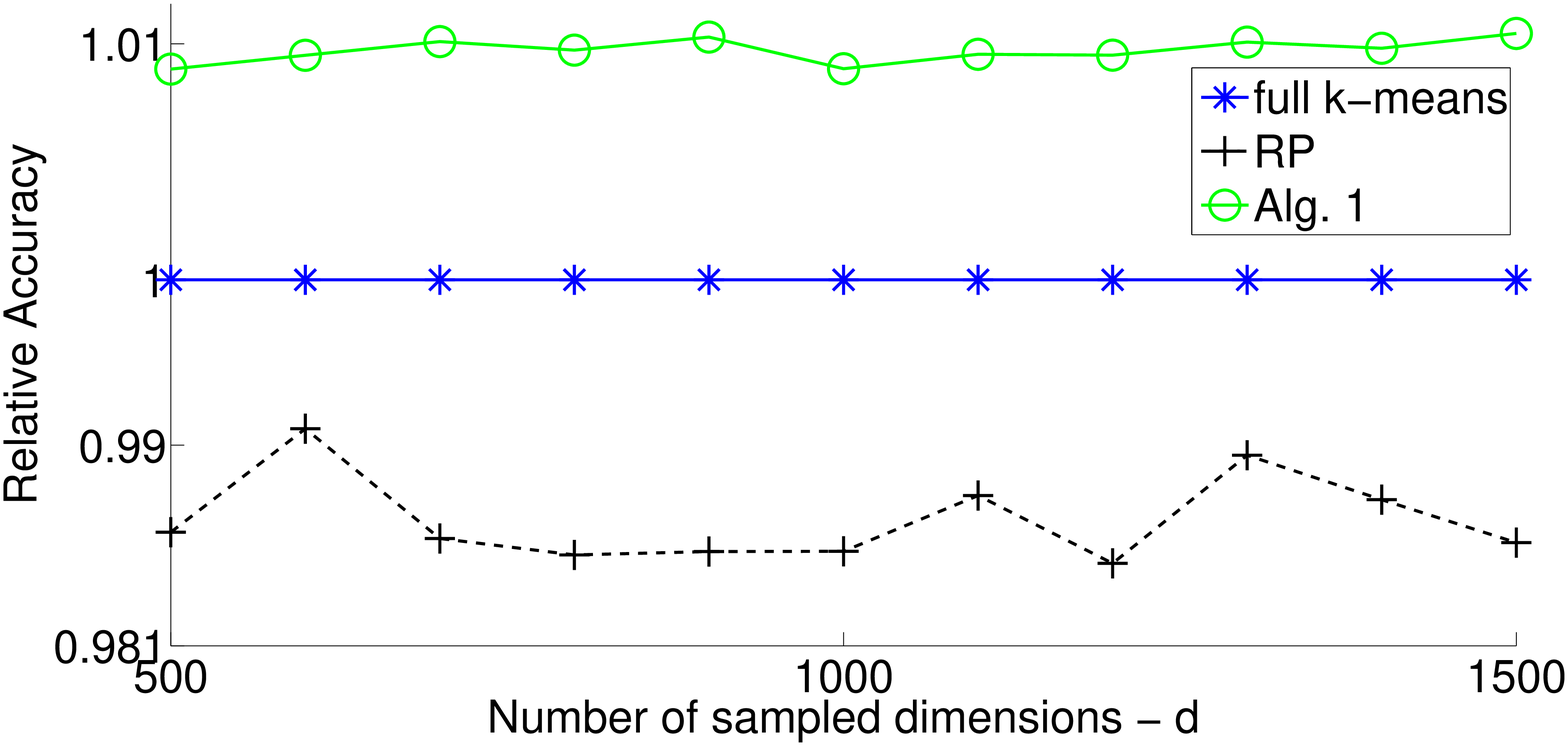}
    \label{fig:kddb_par_acc}}\\
  \centering
  \subfloat[Clustering time (secs)]
  {\includegraphics[width=0.5\columnwidth]{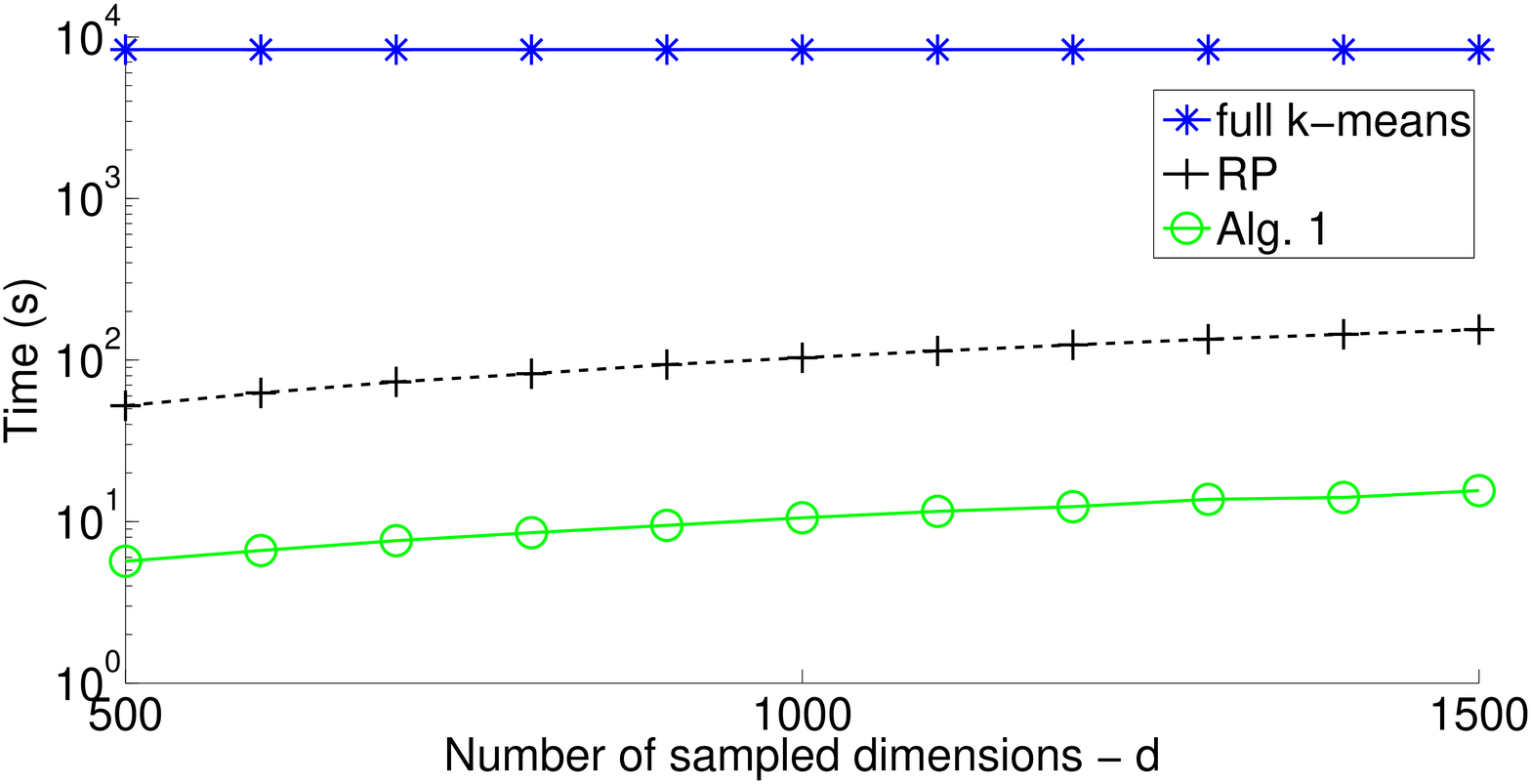}
    \label{fig:kddb_par_time}}
  \caption{A subset of the KDDb data-set with multithreading.}
  \label{fig:kddb_par}
\end{figure}

}

\section{Conclusions and Future Research}

Inspired by RANSAC ideas that have well-appreciated merits for
outlier-resilient regression, this paper introduced a novel
algorithmic framework for clustering massive numbers of
high-dimensional data. Several members of the proposed sketching and
validation (SkeVa) family were introduced. The first two members, a
batch and a sequential one tailored to streaming modes of operation,
required $K$-means clustering in low-dimensional spaces and/or a small
number of data-points. To enable clustering of even nonlinearly
separable data, a third member of the family leveraged the kernel
trick to cluster linearly separable mapped data. A divergence metric
was utilized to develop the fourth member of SkeVa~K-means that
bypasses intermediate $K$-means clustering to trade-off accuracy for
reduced complexity. Extensive numerical tests on synthetic and real
data-sets demonstrated the competitive performance of SkeVa over
state-of-the-art schemes based on random projections. Future research
will focus on the rigorous performance analysis of the proposed
framework, and on the application of SkeVa to spectral clustering and
its MapReduce implementation.

\begin{appendices}
\section{Soft Kernel $K$-means}\label{app:soft.kernel.kmeans}

Since this appendix deals with $N\gg$, the clustering schemes of
Sec.~\ref{sec:prelim} will be applied here to a reduced
number $\check{\nu}$ of $D \times 1$ vectors. Let $\check{\bm{X}}^{(r)}\in
\Real^{D\times\check{\nu}}$ denote the subset of data obtained by
sketching columns of $\bm{X}$. In this context, \eqref{soft.kmeans}
proceeds as follows. For $i = 1,2,\ldots$,
\begin{subequations}\label{kernel.soft.kmeans}
  \begin{align}
    \intertext{[$i$-a] \textbf{Update data-cluster associations:}\ For
      $n=1,\ldots,\check{\nu}$,}
    & \bm{\pi}_n[i] \in \Argmin_{\substack{\bm{\pi}\in [0,1]^K\\ \bm{1}^{\top}\bm{\pi}=1}}
    \delta\left(\varphi\bigl({\bm{x}}_n^{(r)}\bigr),
      \sum\nolimits_{k}\pi_k \bm{c}_k[i]\right) +
    \rho(\bm{\pi})\,.\label{kernel.soft.kmeans.a}\\
    \intertext{[$i$-b] \textbf{Update cluster centroids:}}
    & \{\bm{c}_k[i+1]\}_{k=1}^K \in
    \smashoperator[l]{\Argmin_{\{\bm{c}_k\} \subset\hilbert}}
    \sum_{n=1}^{\check{\nu}} \delta\left(\varphi\bigl({\bm{x}}_n^{(r)}
      \bigr),\sum_{k} \bigl[\bm{\pi}_n[i]\bigr]_k \bm{c}_k
    \right)\,. \label{kernel.soft.kmeans.b}
  \end{align}
\end{subequations}
Recall that given a kernel $\kappa$, if $\varphi(\bm{x}):=
\kappa(\bm{x}, \cdot)$, then inner products in $\hilbert$ can be
obtained as kernel evaluations:
$\innerp{\varphi(\bm{x})}{\varphi(\bm{x}')}_{\mathcal{H}} =
\innerp{\kappa(\bm{x},\cdot)}{\kappa(\bm{x}',\cdot)}_{\mathcal{H}} =
\kappa(\bm{x}, \bm{x}')$, where $\innerp{\cdot}{\cdot}_{\mathcal{H}}$
denotes the inner product in $\mathcal{H}$. Moreover, function
$\delta$ is chosen as
\begin{equation*}
  \delta\left(\varphi\bigl({\bm{x}}_n^{(r)}\bigr),
    \sum\nolimits_{k} [\bm{\pi}_n]_k \bm{c}_k\right) =
  \norm*{\varphi({\bm{x}}_n^{(r)}) - \sum\nolimits_{k} [\bm{\pi}_n]_k
    c_k}_{\hilbert}^2\,,
\end{equation*}
with $\norm{}_{\mathcal{H}} :=
\innerp{\cdot}{\cdot}_{\mathcal{H}}^{1/2}$. It can be shown then by
the Representer's theorem \cite{bishop} that due to the limited number
of data $\{\varphi({\bm{x}}_n^{(r)})\}_{n=1}^{\check{\nu}'}$, looking
for a solution of \eqref{kernel.soft.kmeans.b} in $\mathcal{H}$ is
equivalent to looking for one in the low-dimensional linear subspace
$\hilbert^{(r)} := \linspan\{\varphi(\bm{x}_{1}^{(r)}), \ldots,
\varphi(\bm{x}_{\check{\nu}}^{(r)})\}$, of $\text{rank}\leq
\check{\nu}$, and where $\linspan$ stands for the linear span of a set
of vectors. For notational convenience, let
$\Phi(\check{\bm{X}}^{(r)}):= [\varphi(\bm{x}_{1}^{(r)}), \ldots,
\varphi(\bm{x}_{\check{\nu}}^{(r)})]$, and $\Phi(\check{\bm{X}}^{(r)})
\bm{b} := \sum_{n=1}^{\check{\nu}} b_n\varphi(\bm{x}_n^{(r)})$, for
any $\bm{b}\in \Real^{\check{\nu}}$.

Centroid $\bm{c}_k$ belongs to $\hilbert^{(r)}$, and can be expressed
as a linear superposition of
$\Phi(\check{\bm{X}}^{(r)})$. Specifically, there exists $\bm{b}_k \in
\Real^{\check{\nu}}$ s.t.\ $\bm{c}_k = \Phi(\check{\bm{X}}^{(r)})
\bm{b}_k$. Upon defining $\bm{B} := [\bm{b}_1, \ldots, \bm{b}_K]$,
then $[\bm{c}_1, \ldots, \bm{c}_K] = \Phi(\check{\bm{X}}^{(r)})
\bm{B}$. Moreover,
\begin{equation}
  \delta\left(\varphi\bigl({\bm{x}}_n^{(r)}\bigr),
    \sum\nolimits_{k} [\bm{\pi}_n]_k c_k\right) =
  \norm*{\varphi({\bm{x}}_n^{(r)}) - \Phi\bigl(\check{\bm{X}}^{(r)}\bigr)
    \bm{B}\bm{\pi}_n}_{\hilbert}^2 \nonumber \label{kernel.kmeans.ver.1}
\end{equation}
which can be also obtained through kernel evaluations. Letting
$\bm{K}^{(r)}$ denote the $\check{\nu} \times\check{\nu}$ kernel
matrix with $(n,n')$th entry $[\bm{K}^{(r)}]_{nn'} :=
\kappa({\bm{x}}_{n}^{(r)}, {\bm{x}}_{n'}^{(r)})$, and
$\bm{k}^{(r)}_{n'{}^{(r)}}$ the $\check{\nu}\times 1$ vector with
$n$th entry $[\bm{k}^{(r)}_{n'{}^{(r)}}]_n :=
\kappa({\bm{x}}_{n}^{(r)}, {\bm{x}}_{n'}^{(r)})$, it follows from the
linearity of inner products that for any $\check{\nu}\times 1$ vector
$\bm{\xi}$, $\innerp{\Phi(\check{\bm{X}}^{(r)})
  \bm{\xi}}{\varphi({\bm{x}}_{n}^{(r)})}_{\mathcal{H}} =
\bm{\xi}^{\top}\bm{k}^{(r)}_{n^{(r)}}$ and
$\innerp{\Phi(\check{\bm{X}}^{(r)})\bm{\xi}}{\Phi(\check{\bm{X}}^{(r)})
  \bm{\xi}}_{\mathcal{H}} = \bm{\xi}^{\top}\bm{K}^{(r)} \bm{\xi}$. As
such, the quadratic term in \eqref{kernel.kmeans.ver.1} becomes
\begin{align}
  & \norm*{\varphi\bigl({\bm{x}}_{n}^{(r)}\bigr) -
    \Phi\bigl(\check{\bm{X}}^{(r)}\bigr)
    \bm{B}\bm{\pi}_n}_{\hilbert}^2 \notag\\
  & = \innerp*{\varphi\bigl({\bm{x}}_{n}^{(r)}\bigr)}
  {\varphi\bigl({\bm{x}}_{n}^{(r)}\bigr)}_{\mathcal{H}}
  - 2\innerp*{\Phi\bigl(\check{\bm{X}}^{(r)}
  \bigr)\bm{B}
  \bm{\pi}_n}{\varphi\bigl({\bm{x}}_{n}^{(r)}\bigr)}_{\mathcal{H}} \notag\\
  & \hphantom{=\ } + \innerp*{\Phi
  \bigl(\check{\bm{X}}^{(r)}\bigr)\bm{B}\bm{\pi}_n} {\Phi
  \bigl(\check{\bm{X}}^{(r)}\bigr)\bm{B}\bm{\pi}_n}_{\mathcal{H}} \notag\\
  & = \kappa\bigl({\bm{x}}_{n}^{(r)}, {\bm{x}}_{n}^{(r)}\bigr) - 2
  \bm{\pi}_n^{\top} \bm{B}^{\top} \bm{k}^{(r)}_{n^{(r)}} +
  \bm{\pi}_n^{\top} \bm{B}^{\top} \bm{K}^{(r)}\bm{B}
  \bm{\pi}_n \label{basis.4.kernel.kmeans}
\end{align}
which shows that kernel $K$-means in \eqref{kernel.soft.kmeans} boils
down to solving a finite-dimensional optimization task w.r.t.\
$\bm{B}$ and $\bm{\Pi}:= [\bm{\pi}_1, \ldots,
\bm{\pi}_{\check{\nu}}]$.

Moreover, distances in $\hilbert$ between
$\varphi({\bm{x}}_n^{(r')})\in \Phi(\check{\bm{X}}^{(r')})$ and
centroids $\check{c}_k^{(r)}=\Phi(\check{\bm{X}}^{(r)})
\bm{b}_k^{(r)}$ needed in
step~\ref{alg.kernel.version:cluster.consensus.2.random.samples} of
Alg.~\ref{alg:kernel.SkeVa} can be efficiently computed because
they are also expressible in terms of kernel evaluations as follows;
\begin{align*}
  & \norm*{\varphi \bigl({\bm{x}}_n^{(r')}\bigr) -
    \Phi \bigl(\check{\bm{X}}^{(r)} \bigr)\bm{b}_k^{(r)}}_{\hilbert}^2 \\
  & = \innerp*{\varphi\bigl({\bm{x}}_{n}^{(r')}\bigr)}
  {\varphi\bigl({\bm{x}}_{n}^{(r')}\bigr)}_{\mathcal{H}}
  - 2\innerp*{\Phi\bigl(\check{\bm{X}}^{(r)}\bigr)\bm{b}_k^{(r)}}
  {\varphi\bigl({\bm{x}}_{n}^{(r')}\bigr)}_{\mathcal{H}}\\
  & \hphantom{=\ } + \innerp*{\Phi \bigl(\check{\bm{X}}^{(r)}
    \bigl)\bm{b}_k^{(r)}} {\Phi \bigl(\check{\bm{X}}^{(r)}
    \bigl)\bm{b}_k^{(r)}}_{\mathcal{H}}\\
  & = \kappa \bigl({\bm{x}}_n^{(r')}, {\bm{x}}_n^{(r')}\bigr) - 2
  \bm{b}_k^{(r)}{}^{\top} \bm{k}_{n^{(r')}}^{(r)} + \bm{b}_k^{(r)}{}^{\top}
  \bm{K}^{(r)} \bm{b}_k^{(r)}\,.
\end{align*}
\end{appendices}


\end{document}